\documentclass[final,5p,times,twocolumn]{elsarticle}

\usepackage{amssymb}
\usepackage{amsmath,bm}
\usepackage{algorithm}
\usepackage{algpseudocode}
\usepackage{subcaption} 
\usepackage{multirow}
\usepackage{comment}
\usepackage[colorlinks=true,linkcolor=blue,citecolor=blue,urlcolor=blue]{hyperref}
\usepackage[nameinlink,noabbrev]{cleveref}
\crefname{figure}{Figure}{Figures}
\crefname{table}{Table}{Tables}
\usepackage{graphicx} 
\usepackage{url}

\journal{TBD}

\date{July 2025}

\usepackage{amsfonts}
\usepackage{amsthm}
\usepackage{amsfonts}
\usepackage{amsmath}
\usepackage{amscd}
\usepackage{bbm}
\usepackage{booktabs}
\usepackage{siunitx}
\usepackage[utf8]{inputenc}
\usepackage{t1enc}
\usepackage[mathscr]{eucal}
\usepackage{indentfirst}
\usepackage{graphicx}
\usepackage{adjustbox}
\usepackage{xcolor}
\usepackage{hyperref}
\usepackage{graphics}
\numberwithin{equation}{section}
\usepackage{hyperref}
\usepackage{algorithm}
\usepackage{algpseudocode} 
\usepackage{caption}

\begin{document}

\begin{frontmatter}
 
\title{tBayes-MICE: A Bayesian Approach to Multiple Imputation for Time Series Data}

 \author[1,2]{Amuche Ibenegbu} 
 \author[1,2]{Pierre Lafaye de Micheaux}
 \author[1,2]{Rohitash Chandra}

\affiliation[1]{Transitional Artificial Intelligence Research Group, School of Mathematics and Statistics, UNSW, Sydney, Australia}
\affiliation[2]{School of Mathematics and Statistics, UNSW, Sydney, Australia}

\begin{abstract}

Time-series analysis is often affected by missing data, a common problem across several fields, including healthcare and environmental monitoring. 
 Multiple Imputation by Chained Equations (MICE) has been prominent for imputing missing values through "fully conditional specification". We extend MICE using the Bayesian framework (tBayes-MICE), utilising Bayesian inference to impute missing values via Markov Chain Monte Carlo (MCMC) sampling to account for uncertainty in MICE model parameters and imputed values. We also include temporally informed initialisation and time-lagged features in the model to respect the sequential nature of time-series data. We  evaluate the tBayes-MICE method using two real-world datasets (AirQuality and PhysioNet), and using both the Random Walk Metropolis (RWM) and the Metropolis-Adjusted Langevin Algorithm (MALA) samplers. Our results demonstrate that tBayes-MICE reduces imputation errors relative to the baseline methods over all variables and accounts for uncertainty in the imputation process, thereby providing a more accurate measure of imputation error. 
 We also found that MALA mixed better than RWM across most variables, achieving comparable accuracy while providing more consistent posterior exploration. Overall, these findings suggest that the tBayes-MICE framework represents a practical and efficient approach to time-series imputation, balancing increased accuracy with meaningful quantification of uncertainty in various environmental and clinical settings.

\end{abstract}




\end{frontmatter}

\section{Introduction}

Missing data are prevalent in time-series datasets across many domains, from the environmental sector to clinical recordings \cite{hua2024impact,shadbahr2023impact}. Instrument errors and power challenges can lead to incomplete time-stamped data (missing data) in air quality monitoring, hindering the accuracy of analysis and forecasting  \cite{hua2024impact}. Similarly, physiological time series \textcolor{black}{collected in intensive Care Units (ICUs)} often contain missing observations due to irregular sampling, sensor errors or clinical constraints. If not handled appropriately, these gaps can disrupt data pipelines and significantly degrade the performance of machine learning models \cite{jager2021benchmark}. Missing values can arise due to various reasons (missingness pattern), which can be categorised into three primary cases: Missing Completely At Random (MCAR), Missing At Random (MAR), and Missing Not At Random (MNAR)~\cite{little2019statistical}. MCAR occurs when missingness is independent of both observed and unobserved values, typically due to accidental loss of information. MAR describes cases where missingness depends only on the observed data, and is independent of the missing values themselves, while MNAR refers to non-ignorable missingness that depends directly on unobserved values, often requiring explicit modelling or additional collection, which is often impractical \cite{donders2006gentle,salgado2016missing}.

The two main approaches for handling missing values include either naively ignoring (deletion) or imputing (filling in) them with new values using statistical \cite{velicer2005comparison} and machine learning models \cite{batista2002study, stekhoven2012missforest}. The deletion is suitable only when a small proportion of the data is missing; however, for a significant amount of missing data, data imputation methods provide a more practical and effective solution ~\cite{farhangfar2007novel}. 
The imputation can be categorised into single (e.g. mean, linear interpolation, etc.) and multiple imputation approaches that explicitly account for uncertainty ~\cite{farhangfar2007novel, schafer1999multiple, jager2021benchmark}. 

Multiple Imputation by Chained Equations (MICE)~\cite{van1999flexible} has become a widely used multiple imputation method for multivariate datasets. MICE iteratively fits a sequence of univariate conditional models, imputing each variable with missing values using the remaining variables as predictors ~\cite{azur2011multiple}. This procedure accounts for imputation uncertainty by producing multiple completed datasets using different random seeds ~\cite{erler2019bayesian,hua2024impact}.
MICE is popular because it is flexible, easy to apply, and, under the MAR assumption, tends to yield parameter estimates close to unbiased. This iterative mechanism is also known as a \emph{Fully Conditional Specification} (FCS) framework ~\cite{van2011mice}, and the updates performed by MICE can be viewed as \emph{Gibbs-style} iterations, in which variables are successively imputed from their estimated conditionals until convergence.

Regardless of its popularity, standard MICE is not designed for time-series settings.   It assumes by default that observations are interchangeable and ignores temporal ordering unless time indices or lagged variables are added as predictors ~\cite{dudhe2025icu}. When applied to time series data, this assumption can lead to unrealistic imputations that break inherent temporal behaviour,  such as trends or seasonality. This can lead to implausible imputations that disrupt time-dependent patterns, such as trends or seasonality. Although the MICE approach is inspired by Bayesian ideas, it does not represent a full Bayesian framework. During each round of the MICE imputation process, the model parameters are typically treated as being fixed rather than random. This is because there is no formal Bayesian propagation of uncertainty about the model parameters at each imputation round. In addition, a full joint probability density function may not exist if the conditional models used in the MICE algorithm are not mutually compatible.  This can lead to biased posterior draws and an underestimation of the between-imputation variance, meaning that the true uncertainty associated with the missing data will not be reflected in the analysis results ~\cite{murray2018multiple,erler2019bayesian}. 
Recently developed alternatives to the imputation framework include learning-based extensions of the MICE algorithm \cite{naf2024good}. This work focused on cross-sectional datasets and has only briefly addressed issues related to temporal dependence and uncertainty propagation. An alternative approach to the use of learning-based extensions of MICE is to incorporate Markov Chain Monte Carlo (MCMC) methods into the imputation process \cite{gilks1995markov, robert1999monte, rubin1978multiple}. In addition to allowing model parameters to be updated probabilistically, when using MCMC in conjunction with imputation, the missing values are also updated probabilistically.

These challenges highlight the need for imputation frameworks that can respect temporal structure and accurately account for data uncertainty. In this study, we introduce \emph{tBayes-MICE}, a time-aware Bayesian extension of MICE that integrates MCMC sampling within the chained equation framework. It features temporally informed initialisation strategies and time-lagged features that produce imputations drawn from richer posterior distributions, thereby improving coherence in multivariate time-series settings. We implemented two MCMC samplers, the Random Walk Metropolis (RWM) and the Metropolis-Adjusted Langevin Algorithm (MALA), and evaluated their performance on the AirQuality and clinical PhysioNet datasets.

This paper is organised as follows. In Section~2, related work on data imputation for time series, MICE, and MCMC is presented.
Section~3 outlines the methodology, including Data description, MICE, and tBayes-MICE. Sections~4 and~5 present the results and
Discussion respectively, and Section~6 concludes the study.

\section{Related Work}

\subsection{Time Series Data Imputation}

Past studies have evaluated numerous approaches to handling missing data in time series, including both traditional statistical methods and modern machine learning and deep learning models. Most traditional statistical methods utilise temporal information to predict missing values. For instance, State-Space Models with Kalman Filter can naturally treat missing values as "latent" states. As demonstrated by Moritz et al.~\cite{moritz2015comparison}, Kalman Smoothing with Structural Time-Series Models is among the best-performing univariate imputation methods, while Velicer and Colby \cite{velicer2005comparison} found that maximum likelihood estimation in ARIMA models does substantially better than simply substituting the mean value of the corresponding variable for the missing value. Yamoah et al.~\cite{afrifa2020missing} compared several imputation methods on an hour-based dataset and determined that Multiple Linear Regression and ARIMA-Kalman Smoothing were able to produce reasonably accurate imputed values.   Because they are easy to apply and relatively fast, smoothing, interpolation \cite{noor2015comparison}, and spline methods \cite{junninen2004methods} are widely applied; however, they rarely capture complex temporal interactions and relationships between variables \cite{che2018recurrent}. Therefore, we use a few of these older methods as a baseline for evaluating the univariate MICE method in this study.

In recent years, Machine Learning Models have been extensively utilised for Data Imputation. In early studies,  Batista and Monard \cite{batista2002study} demonstrated that K-Nearest Neighbours (KNN) was able to perform better than the inherent Missing-Value Handling Mechanisms in Decision Tree-based Classifiers. Ensemble Learning Methods such as missForest \cite{stekhoven2012missforest} further improved upon the performance of earlier models by utilising Random Forests to capture non-linear relationships between variables. More recently, data imputation has been viewed from a predictive perspective.  Morvan et al.~\cite{le2021sa} demonstrated that impute-then-regress strategies can be theoretically optimal for supervised learning and proposed the NeuMiss method, which integrates imputation into the prediction pipeline. However, the authors did not focus on uncertainty propagation and temporal coherence, which are central to time-series analysis.

In the last decade, Deep Learning models have made substantial contributions to improving time series imputation performance, especially with Recurrent Neural Networks (RNNs). Che et al.~\cite{che2018recurrent} introduced the GRU-D model, which directly incorporates the missingness of input data and time gaps between observations into its architecture, allowing the network to learn informative missing patterns. The BRITS model \cite{cao2018brits} extended the previous model by employing a bidirectional RNN architecture and treating missing values as trainable parameters, yielding strong performance on real datasets. Some subsequent models, such as CATSI \cite{yin2020context}, employed a context-aware mechanism to learn the global temporal structure and, more importantly, to capture the temporal dependencies. Bayes-CATSI \cite{kulkarni2024bayes} extends CATSI by adding Bayesian components to enhance the robustness and provide uncertainty estimates, particularly for clinical data.

\subsection{MICE}

 MICE \cite{van1999flexible,van2011mice} is one of the most commonly used methods for handling missing data within the Fully Conditional Specification (FCS) framework. It iteratively imputes missing values for each variable using a regression model conditioned on all other variables. MICE is flexible enough to accommodate a wide range of variables and model structures, and it introduces randomness to capture the uncertainty associated with the imputation process. Several variations of MICE have been developed to improve its efficiency and robustness. In this regard, Brand \cite{brand1999development} developed a version of MICE using Gibbs sampling that includes only the most important predictors for each variable, resulting in simpler models and improved convergence.
Khan and Hoque \cite{khan2020sice}, inspired by MICE, developed a new hybrid approach, known as the Single Centre Imputation from Multiple Chained Equations (SICE), that combines single and multiple imputation to handle both numerical and categorical variables. More recently, MICE has been extended to handle more complex data structures. Rigon and White \cite{resche2018multiple}, for example, developed a multi-level MICE framework that performs well even when cluster-level predictors are omitted. 

Naf et al.~\cite{naf2024good} addressed the theoretical and practical issues of MICE further by evaluating what makes an effective imputation strategy in terms of robustness under the MAR mechanism. The authors proposed MICE Distributional Random Forest (mice-DRF) as a means to effectively capture the non-linear relationship and distributional changes of covariates and demonstrated the shortcomings of using RMSE metrics as a measure of uncertainty; they then presented the m-I score as a metric that measures the effectiveness of imputed values to approximate the Full Conditional Distribution. Further, Grzesiak et al. \cite{grzesiak2025we} conducted large-scale benchmarking and found that FCS methods, especially MICE, generally preserve data distributions much better than joint modelling and Deep Learning methods. Nevertheless, most MICE variants remain deterministic and do not effectively propagate parameter uncertainty, motivating Bayesian extensions of MICE.

\subsection{MCMC sampling}
MCMC sampling is a family of iterative computational methods used to implement Bayesian inference and to sample from the posterior distribution of unknown model parameters or missing values in the observed data. Gibbs sampling \cite{geman1984stochastic,gelfand1990sampling} is a type of MCMC sampling algorithm that generates samples iteratively from conditional distributions and can therefore be used for imputing missing values when the joint distribution of the variables is complex \cite{van2018flexible}. The Data Augmentation algorithm for Bayesian inference with missing data was first introduced by Tanner and Wong \cite{tanner1987calculation}, and it has served as the foundation for all modern MCMC-based imputation techniques. Schafer \cite{schafer1997analysis} implemented Gibbs sampling for practical use in multivariate normal and general location models. In addition to the Gibbs Sampler, several other MCMC variants can be used to improve sampling efficiency in high-dimensional or structured settings. Metropolis et al.~\cite{metropolis1953equation} developed the Random Walk Metropolis (RWM), which generates samples by proposing new states with Gaussian noise added to the current state. This allows the algorithm to explore the parameter space through a random walk. Later, Hastings generalised this idea \cite{hastings1970monte} and proposed the Metropolis-Hastings (MH) algorithm, which allows the use of a wide range of proposal distributions. For even higher efficiency, gradient-based samplers like the Metropolis-Adjusted Langevin Algorithm (MALA) \cite{roberts1996exponential} make use of the geometric properties of the posterior surface to improve the mixing of the samples by combining the information contained in the gradients with the information contained in the noise.

All these improvements can be very useful when dealing with time-series and spatial imputation, since in both contexts, the posterior surfaces are often very informative but difficult to explore. However, the theoretical results have provided important guidelines for tuning MCMC algorithms. The diffusion limit theory, studied by Robert et al. \cite{gelman1997weak}, has also provided theoretical results on optimal scaling; in particular, they found that the optimal acceptance rate is approximately 23\% for RWM in high-dimensional cases. Subsequently, Roberts and Rosenthal \cite{roberts1998optimal, roberts2001optimal} have extended the framework of optimal scaling to a large number of Metropolis-Hasting algorithms, including MALA. Their work has demonstrated how tuning strategies can be applied to different variants of algorithms and how crucial the selection of the proposal's scale is for efficient exploration of the posterior space. Several empirical studies have shown that integrating MCMC-based methods into the multiple imputation workflow can make valuable contributions to real-world problems. As an example, Yozgatligil et al.~\cite{yozgatligil2013comparison} have shown how the Monte Carlo Markov Chain based on expectation–maximization (EM-MCMC) algorithm outperforms the traditional EM algorithm for the imputation of Turkish meteorological time series data, while Sun \cite{sun2017application} has shown that the Gibbs-based multiple imputation technique is highly effective in clinical datasets with complex missingness mechanisms.

\section{Methodology}

We demonstrate the effectiveness of our proposed framework using two distinct real-world datasets and equally present the algorithms of MICE and tBayes-MICE. 

\subsection{Data Description}

\subsubsection{Air Quality Data}

The dataset consists of 9,358 hourly averaged records across 15 variables, collected between March 2004 and February 2005 using a chemical multisensor device deployed at road level in a highly polluted urban area in Italy. The device captured signals from five metal oxide sensors, while ground truth concentrations of key pollutants, Carbon Monoxide (CO), Non-Methane Hydrocarbons (NMHC), Benzene (C\textsubscript{6}H\textsubscript{6}), Nitrogen Oxides (NO$_\text{x}$), and Nitrogen Dioxide (NO$_\text{2}$) were measured using a certified co-located analyser. We utilised environmental variables, including temperature and humidity, from this dataset. De Vito et al.~\cite{de2008field} used the data for urban pollution monitoring and highlighted challenges such as sensor drift and cross-sensitivities. 

For our study, we randomly selected eight features: Date, Time, CO (GT), PT08.S1 (CO), NMHC (GT), C\textsubscript{6}H\textsubscript{6} (GT), PT08.S2 (NMHC), and Temperature (T). To enable time-aware initialisation in our imputation strategy, we merged the Date and Time columns into a single timestamp variable (Date-Time). We denoted missing values as -200 and, after filtering out all rows containing them, retained 863 complete observations. We then introduced 20\% artificial missingness under the assumption of  MCAR to evaluate imputation performance under controlled conditions.

\subsubsection{Healthcare Data}

We used set A from the 2012 PhysioNet/Computing in Cardiology challenge, which comprises clinical time-series data from 4,000 intensive care unit (ICU) patients \cite{silva2012predicting}. Each patient's data is stored in an individual ASCII file, which may contain up to 48 hours of irregularly sampled physiological and laboratory data. These include vital signs such as heart rate, respiratory rate, sodium (Na\textsuperscript{+}), white blood cell count (WBC), and other important ICU monitoring parameters.

We converted the raw time-series data into a tabular format to enable structured analysis and imputation. Since each row in the table represented a unique patient and each column represented a time-averaged parameter, we developed a custom Python pipeline to uniformly extract both static and dynamic features from the individual files. We randomly chose eight parameters for this study: Record ID, Time, Glucose, Bicarbonate (HCO$_\text{3}$), Magnesium (Mg), Na\textsuperscript{+}, Platelet, and WBC. Prior to developing the current dataset, many patient-time records had a large number of missing values, with some rows exceeding 60\%. To filter the data and improve data quality by reducing the effect of extreme sparseness, we removed all rows with more than 60\% missingness. After this filtering step, we obtained a refined dataset of 3,774 patient-time records, with an overall missing rate of 2.87\% for the selected parameters.

We first removed all naturally occurring missing values (2.87\%), to assess how well our imputation methods would perform under data collection conditions similar to those found in real-world clinical applications. After removing all naturally occurring missing values, we were left with a completely observed subset of the data. We then added a structured pattern of missing data to the completely observed data, using a MissingnessPattern class. More specifically, we added a MAR pattern by masking approximately 70\% of the variables (i.e., 4 out of 6 physiological variables), leaving the remaining 30\% unmasked (fully observed). We implemented the MAR process using a logistic regression-based masking approach, in which linear combinations of observed variables determine the probability of missingness for each variable. We assigned each masked variable a 40\% probability of missingness, yielding an overall induced missingness rate of approximately 26.7\%.

While the induced missingness rate is significantly higher than the natural missingness rate of 2.87\%, it was deliberately chosen to test the imputation frameworks' ability under more difficult, heterogeneous missingness conditions. In clinical applications, missing data are often unevenly distributed and can be particularly pronounced in specific patient populations or measurement contexts (e.g., unstable patients or infrequently measured biomarkers). By inducing a structured yet high level of missingness in the data, we evaluate the robustness, reliability, and generality of the imputation models and demonstrate that they can function appropriately across a wide variety of imperfect real-world clinical settings.

\subsection{Models}

\subsubsection{Time Series MICE}

Multiple Imputation \cite{rubin1988overview} provides a framework for handling missing values by accounting for uncertainty across multiple complete datasets. Each missing entry is replaced with $m$ plausible values generated by an imputation model, and the final inference combines results across these datasets using Rubin's rule. 
MICE is a widely used approach that implements multiple imputation by iteratively drawing missing values from conditional models using an FCS approach \cite{van2011mice}. This approach imputes each variable sequentially (one at a time) using its conditional distribution, also referred to as regression switching or variable-by-variable imputation \cite{van1999multiple}. 

Let $\boldsymbol{X} = (X_1, X_2,\ldots, X_p)^\top$ be a vector of $p\geq 1$ random variables observed over $n$ time points. Some of these observations might be missing for some of the $p$ variables. The resulting incomplete dataset with time-indexed observations is stored in a rectangular table $\mathbf{X}^0=(x_{t,j})=[\boldsymbol{x}_1\,\cdots\,\boldsymbol{x}_p]$ of size $n\times p$. For a given variable $X_j$ that has missing values, we treat $X_j$ as a response variable and $\boldsymbol{X}_{-j}$ as the vector of remaining variables used as predictors in the conditional imputation model. 
Each random variable $X_j$ ($j=1,\ldots,p$) may be partially observed, so that the vector of its $n$ values in the $j$-th column of $\mathbf{X}^0$, i.e., $\boldsymbol{x}_j=(x_{1j},\ldots,x_{n,j})^\top$, contains elements (indexed by $I_j^{\mathrm{obs}}$) with a value that has actually been observed, while the other elements (indexed by $I_j^{\mathrm{miss}}$) correspond to missing values, encoded as \textrm{NA} (Not Available). Therefore, we have $\{1,\ldots,n\}=I_j^{\mathrm{obs}}\cup I_j^{\mathrm{miss}}$ with $I_j^{\mathrm{obs}}\cap I_j^{\mathrm{miss}}=\emptyset$. We also define $J^{\mathrm{miss}}\subset\{1,\ldots,p\}$ as the set of all columns in $\mathbf{X}^0$ that contain at least one missing value.

In a time-series setting, we observe each variable sequentially, for $t=1,\ldots,n$, to account for temporal dependence within the FCS framework. We denote by $X_{t,j}$ the $j$th random variable at time $t$. To retain the  FCS structure of MICE, temporal dependence is introduced by augmenting the predictor set $\mathbf{X}_{-j}^0$ with lagged observations of the variable $X_j$. Specifically, we define as a temporal predictor of $X_j$ the $((p-1)+\ell_p+\ell_f)\times1$ vector $\boldsymbol{Z}_{-j}$ that complements $\boldsymbol{X}_{-j}$ with $\ell_p+\ell_f$ additional lagged variables defined from $X_j$. A time $t$, we thus have
\[
\boldsymbol{Z}_{t,-j} 
=
\big(\boldsymbol{X}_{t,-j},
X_{t-\ell_p,j}, \ldots, X_{t-1,j}, X_{t+1,j}, \ldots, X_{t+\ell_f,j}
\big)^\top,
\]
where $\ell_p$ and $\ell_f$ denote the number of past and future lags, respectively, and only available observations are used. This is allowed because the objective is imputation rather than forecasting (prediction)\cite{van2018flexible}, and conditioning on both past and future observations makes sense here.

Thereafter, $f_j$ denotes a regression model (e.g., linear regression for quantitative responses, or logistic regression for binary responses). 
Conditioning on the lagged predictors, $f_j$ defines the conditional mean
\[
\mathbb{E}(X_j \mid \boldsymbol{Z}_{t,-j})
= f_j(\boldsymbol{Z}_{t,-j}),
\] 
which partially specifies the full conditional distribution
\[
\mathbb{P}\!\left(
X_j \mid \boldsymbol{Z}_{t,-j} =\boldsymbol{z}_{t,-j} 
\right).
\]
We denote the estimators of both quantities  by $\widehat{f}_j$ and 
$\widehat{\mathbb{P}}_j(\boldsymbol{z}_{t,-j})$,  the latter corresponds to either an empirical or a model-based conditional distribution. 

\begin{figure}
    \centering
    \includegraphics[width=\linewidth]{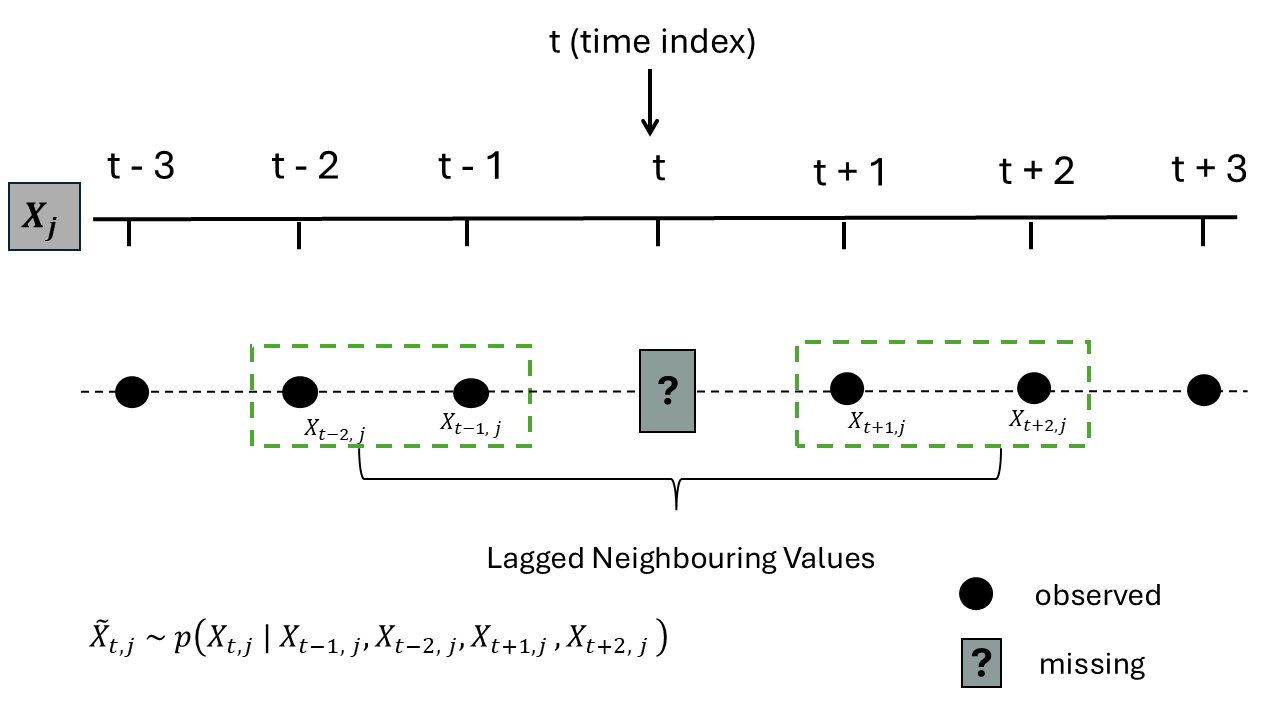}
    \caption{Time-lagged imputation mechanism in univariate MICE, where a missing value at time index $t$ is imputed by conditioning on observed neighbouring values from past and future lags.}
    \label{fig:MICE_Framework}
\end{figure}

In the univariate time-series setting ($p=1$), observations correspond to a single variable indexed over time. As illustrated in \cref{fig:MICE_Framework}, the predictor set is constructed using time-lagged values of the same variable, given by
\[
\boldsymbol{Z}_{t,-j} 
=
\big(X_{t-\ell_p,j}, \ldots, X_{t-1,j}, X_{t+1,j}, \ldots, X_{t+\ell_f,j}
\big).
\]
Assuming that the data are MAR, we present the time-lagged FCS procedure in Algorithm~1. 

\begin{algorithm}[H]
\caption{Time Series MICE}
\begin{algorithmic}[1]
\State \textbf{Input:} Incomplete original dataset $\mathbf{X}^0$ with $p$ time-indexed variables, lag orders $(\ell_p,\ell_f)$, from which we derive the index sets $J^{\mathrm{miss}}$ and, for each $j\in J^{\mathrm{miss}}$, the subsets $I_j^{\mathrm{obs}}$ and~$I_j^{\mathrm{miss}}$. Number of iterations~$K$.
\State \textbf{Step 1:} Create a copy of $\mathbf{X}^0$, denoted by $\mathbf{X}$, which will serve as a placeholder matrix to store the intermediate and final imputed values.
Initialise $\mathbf{X}$ by imputing its missing entries with crude estimates (e.g., the column mean or median) to obtain a complete dataset. 
\State \textbf{Step 2:} Iteratively update the missing entries of $\mathbf{X}$ as follows:
\For{$k = 1$ to $K$}
    \For{each $j\in J^{\mathrm{miss}}$}
    \State $\bullet$ At time index $t$, construct lagged predictors 
    $\boldsymbol{Z}_{t,-j}$ 
    \State from $\mathbf{X}$, excluding the current time point.
    \State $\bullet$ Fit a regression model $f_j$ relating $X_j$ to $\boldsymbol{Z}_{t,-j}$ using
    \State only the observations in rows $I_j^{\mathrm{obs}}$ of $\mathbf{X}$.
    \State $\bullet$ For each $t \in I_j^{\mathrm{miss}}$, draw a predicted value $\widehat{x}_{t,j}$ from 
        \State the predictive distribution $\widehat{\mathbb{P}}_j(\boldsymbol{z}_{t,-j})$, where $\boldsymbol{z}_{t,-j}$ is 
        \State  taken from $\mathbf{X}$, and impute it 
        in place at the position 
        \State $(t,j)$ of $\mathbf{X}$.
    \EndFor
\EndFor    
\State Store $\mathbf{X}$.
\State \textbf{Step 3:} Repeat \textbf{Step~2} $m$ times with different random seeds (and starting each time with the complete dataset from \textbf{Step~1}) to generate $m$ imputed datasets $\mathbf{X}^1,\ldots,\mathbf{X}^m$.
\State \textbf{Output:} The $m$ completed datasets.
\end{algorithmic}    
\end{algorithm}

We describe the main steps of Algorithm 1  in detail to clarify its implementation.

\paragraph{Step 1 (Initialisation)} 
We initially impute all missing entries in $\mathbf{X}^0$  with simple deterministic values \textcolor{black}{(e.g., column means)} to obtain a complete dataset~$\mathbf{X}$. This step does not produce accurate imputations but rather provides a coherent starting point for constructing lagged predictors. In particular, initialising the series ensures that past and future lag values required by the time-lagged conditional models are available at the beginning of the iterative procedure.

\paragraph{Step 2 (Iterative imputation)}

For each iteration $k = 1,\ldots,K$, we consider the time series variable $X_j$ as the response variable at time index $t$ where observations are missing. 
Instead of conditioning solely on other variables, as in classical multivariate MICE, the predictor vector $\boldsymbol{Z}_{t,-j}$ (model input) is constructed using lagged observations of $X_j$, optionally augmented with contemporaneous values of the remaining variables $\boldsymbol{X}_{-j}$. Specifically, for each missing time point $t \in I_j^{\mathrm{miss}}$, we then fit a regression model $f_j$ using the observed time points $\{(x_{t,j},\boldsymbol{z}_{t,-j}) : t\in I_j^{\mathrm{obs}}\}$ where
\[\boldsymbol{Z}_{t,-j} =
\big(\boldsymbol{X}_{t,-j},X_{t-\ell_p,j}, \ldots, X_{t-1,j}, X_{t+1,j}, \ldots, X_{t+\ell_f,j}
\big)\]  for the general case ($p>1$), and 
\[\boldsymbol{Z}_{t,-j} =
\big(X_{t-\ell_p,j}, \ldots, X_{t-1,j}, X_{t+1,j}, \ldots, X_{t+\ell_f,j}
\big)\] for the univariate case ($p=1$).
This model defines the estimated conditional mean 
$\widehat{f}_j(\boldsymbol{z}_{t,-j}) = \mathbb{E}(X_j\mid\boldsymbol{Z}_{t,-j} =\boldsymbol{z}_{t,-j})$, 
and the associated estimated conditional distribution 
$\widehat{\mathbb{P}}(X_j\mid\boldsymbol{Z}_{t,-j} =\boldsymbol{z}_{t,-j})$, which may be parametric or empirical.
For each $i\in I_j^{\mathrm{miss}}$, a random draw
\[
\widehat{x}_{t,j} \sim 
\widehat{\mathbb{P}}\!\big(X_j\mid\boldsymbol{Z}_{t,-j} =\boldsymbol{z}_{t,-j})
\]
is taken, where $\boldsymbol{z}_{t,-j}$ is extracted from $\mathbf{X}$. This value $\widehat{x}_{t,j}$ is inserted (in-place) into $\mathbf{X}$. 
 The algorithm propagates information through time via lagged dependencies by iteratively updating the series in this manner. Although no explicit joint distribution is specified, the repeated conditional sampling across iterations yields behaviour similar to a Gibbs sampler on temporally structured conditionals.

\paragraph{Step 3 (Multiple imputations)}
Step~2 is repeated $m$ times with different random seeds to generate 
$m$ completed datasets, each representing a possible realisation of the missing values under the MAR assumption.

\subsubsection{tBayes-MICE}

We introduce \emph{Bayesian MICE} (tBayes-MICE), a Bayesian extension of the classical MICE framework for multivariate time series data using MCMC sampling. tBayes-MICE retains the  FCS structure of MICE, but replaces deterministic regression updates with Bayesian regression models whose parameters and imputations are jointly sampled using MCMC. This formulation enables uncertainties to be propagated through both model parameters and imputed values, yielding probabilistically coherent imputations.


Let $\boldsymbol{X} = (X_1, X_2,\ldots, X_p)$ denote  a vector of $p\geq 1$ variables observed sequentially over time, measured on $n$ individuals, with the data stored in a rectangular matrix 
$\mathbf{X}=(x_{t,j})=[\boldsymbol{x}_1\,\cdots\,\boldsymbol{x}_p] \in \mathcal{R}^{n \times p}$, where $t=1, \ldots, n$ indexes time. Each variable $X_j$ may be partially observed, with observed and missing index sets denoted by $I_j^{\mathrm{obs}}$  and $I_j^{\mathrm{miss}}$ respectively. The set of variables containing at least one missing value is denoted by $J^{\mathrm{miss}}\subset\{1,\ldots,p\}$.
As in classical MICE, tBayes-MICE iteratively imputes each variable $X_j$ conditional on the others. 
However, to account for temporal dependence, we augment the conditional models with lagged information (past and future or just past) and time features. 

At time index $t$, the predictor vector for imputing $X_j$ is defined as 
\[\boldsymbol{Z}_{t,-j} = \big( X_{t,-j},\; h_t,\; X_{t\pm 1:t\pm \ell,j} \big)\]

\[\boldsymbol{Z}_{t, -j}=\Big(X_{t,-j},\;
h_t,\;X_{t-1,j}, \ldots, X_{t-\ell,j},\;
X_{t+1,j}, \ldots, X_{t+\ell,j}\Big),\]

where $X_{t,-j}$ denotes the contemporaneous values of all predictors at a time $t$, except $X_j$, $h_t$ represents time-derived covariates (e.g., hour-of-day, and day-of-week), $\ell$ is the maximum lag order. Since it is an imputation and not forecasting \cite{van1999flexible}, we used bidirectional lag for $X_j$ instead of strictly forward-looking, with a maximum lag of 2. 
For each variable $X_j$, we specify a Bayesian regression model 
\begin{equation}
X_{t,j} = \boldsymbol{Z}_{t,-j}^\top \theta_j + \varepsilon_{t,j},
\qquad
\varepsilon_{t,j} \sim \mathcal{N}(0,\tau_j^2)
\end{equation}

\begin{equation}
X_{t,j} \mid \boldsymbol{Z}_{t,-j}, \theta_j, \tau_j^2
\sim \mathcal{N}\!\left(\boldsymbol{Z}_{t,-j}^\top \theta_j,\; \tau_j^2\right)
\end{equation}
where $\theta_j$ denotes the regression coefficients (including the intercept), and $\tau_j^2$ is the residual variance.

The likelihood for the observed entries is given by

\begin{equation}
\mathcal{P}(X_j^{\mathrm{obs}} \mid \boldsymbol{Z}_{-j}, \theta_j, \tau_j^2)
=
\frac{1}{(2\pi\tau_j^2)^{\frac{n_j}{2}}}
\exp\!\left(
-\frac{1}{2\tau_j^2}
\sum_{t \in I_j^{\mathrm{obs}}}
\big(
x_{t,j} - \boldsymbol{Z}_{t,-j}^\top \theta_j
\big)^2
\right).
\end{equation}

 These \textcolor{black}{regression parameters} are treated as random variables, allowing us to capture estimation uncertainty. We assumed a multivariate Gaussian prior for $\theta_j$ and a non-negative prior for $\tau^2$, which is why we assume an inverse-Gamma distribution as done in earlier studies \cite{chandra2024bayesian}. Therefore, we use weakly informative priors to guide this process without being overly restrictive:
\[\theta_j \sim \mathcal{N}(0, \sigma^2 I), \quad  \tau_j^2 \sim \mathcal{IG}(v_1, v_2)\]
These priors ensure that the data can strongly influence the posterior while preventing extreme or implausible values. 
\textcolor{black}{In Bayesian theory, the posteriors are sampled from the combination of the likelihood and the prior and inference is performed in log-posterior space. For each variable $X_j$, we obtain the posterior distribution of the regression coefficients $\theta_j$ and variance parameter $\tau^2$ by combining the Gaussian log-likelihood of the observed data with the log-prior distributions. The resulting log-posterior is given as}
\begin{equation}
\begin{aligned}
\log \mathcal{P}(\theta_j, \tau_j^2 \mid X_j^{\mathrm{obs}}, \boldsymbol{Z}_{-j})
&=
\log \mathcal{P}(X_j^{\mathrm{obs}} \mid \theta_j, \tau_j^2, \boldsymbol{Z}_{-j}) \\
&\quad + \log \mathcal{P}(\theta_j)
+ \log \mathcal{P}(\tau_j^2)
\end{aligned}
\end{equation}

which is equivalent to
\begin{equation}
\log p\!\left(X_j^{\mathrm{obs}} \mid \theta_j, \tau_j^2, \boldsymbol{Z}_{-j}\right)
=
\sum_{t \in I_j^{\mathrm{obs}}}
\log p\!\left(X_{j,t} \mid \theta_j, \tau_j^2, \boldsymbol{Z}_{t,-j}\right)
\end{equation}

We obtain the posterior samples of the parameters $\theta_j^\text{(s)}$ and $\tau_j^\text{2(s)}$ at iteration $s$   using two MCMC samplers, including Random Walk Metropolis (RWM) and the Metropolis-Adjusted Langevin Algorithm (MALA). Given parameter draws at iteration $s$, missing values are imputed from the posterior predictive distribution:
\[
\tilde{x}_{t,j}^{(s)} \sim f_{j}\!\left(Z_{t,-j}^{(s)},\, \theta_{j}^{(s)}\right), \qquad t \in I_{j}^{\mathrm{miss}}
\]

which equivalently corresponds to:
\[
X_{t,j}^{\mathrm{mis}} \sim
\mathcal{N}\!\left(
\boldsymbol{Z}_{t,-j}^{\top} \theta_j^{(s)},\;
\tau_j^{2(s)}\right),\qquad t \in I_j^{\mathrm{miss}}.\]

This formulation ensures that both parameter and predictive uncertainty are coherently incorporated into the imputation process, resulting in more robust and probabilistically valid imputations.

Before the MCMC sampling process,  we initialise the missing values just as in standard MICE, either using mean-based initialisation, or  time-aware initialisation exploiting trend and seasonality (e.g., STL decomposition) for more informed starting values. 

Finally, we apply optimal scaling principles to fine-tune the proposal distributions for both RWM and MALA. For the RWM sampler, we follow the guidelines from the literature \cite{gelman1996efficient,roberts2001optimal, haario2001adaptive}, and adapt the proposal covariance matrix based on the dimensionality of the parameter vector. Given that sufficient data are available (i.e., greater than $d + 5$ observations), we use the empirical covariance structure of the predictors: 
\begin{equation}
    \Sigma_{\text{proposal}} = \frac{2.38^2}{d} \cdot \Sigma_{\text{empirical}}  + \epsilon I_d 
\end{equation}

We fall back to a scaled identity matrix(i.e., fewer than $d + 5$) when the data is limited:
\begin{equation}
   \Sigma_{\text{proposal}} = \frac{2.38^2}{d} \cdot  I_d 
\end{equation}

$\Sigma_{\text{empirical}}$ is the empirical covariance of the design matrix constructed from predictors $X_{-j}$, including a bias term.
$\epsilon > 0$ is a small regularisation constant (e.g., $\epsilon = 10^{-6}$) to ensure positive-definiteness.
 $d$ is the dimensionality of the regression model (including bias).
$I_d$ is the identity matrix in $\mathbb{R}^{d*d}$.

For the MALA sampler, we adopt the optimal-scaling rule by Roberts and Rosenthal \cite{roberts1998optimal}, which shows that step size should scale with the parameter dimension as:
\begin{equation}
    \epsilon_{\mathrm{MALA}} = \frac{1.65^2}{d^{1/3}} 
\end{equation}

which corresponds to an isotropic proposal covariance 

\begin{equation}
    \Sigma_{\text{proposal}} = \epsilon_{\text{MALA}}^2 \cdot  I_d 
\end{equation}

$\epsilon_{\mathrm{MALA}}$ is the step size and $d$ remains the parameter vector. The noise variance parameter $\tau^2$ is sampled via Gibbs updates.
This approach aligns with the adaptive Metropolis method of Haario et al. \cite{haario2001adaptive}, which adapts proposals based on the empirical data structure. This ensures that our MCMC samplers explore the parameter space efficiently and maintain good mixing properties, typically aiming for acceptance rates of around 23.4\% for RWM and 57.4\% for MALA. We provide details of this procedure in Algorithm 2.

\begin{algorithm}[htbp!]
\caption{tBayes-MICE for Multivariate Time Series with MCMC sampling.}
\begin{algorithmic}[1]
\State \textbf{Input:}Incomplete time-series dataset $\mathbf{X}_0 \in \mathbb{R}^{n \times p}$; lag orders $(\ell_p,\ell_f)$; number of MICE iterations $K$; number of imputations $m$. Derive index sets $J^{\mathrm{miss}}$ and $\{I_j^{\mathrm{obs}}, I_j^{\mathrm{miss}}\}_{j \in J^{\mathrm{miss}}}$.

\State \textbf{Step 1:} Create a copy of $\mathbf{X}_0$, denoted by $\mathbf{X}$, which will serve as a placeholder matrix to store the intermediate and final imputed values. Initialise $\mathbf{X}$ by imputing its missing entries with crude estimates (mean or time-aware values).
\State \textbf{Step 2:} Iteratively update the missing entries of $\mathbf{X}$ as follows:
\For{iteration $k = 1$ to $K$}
\For{each $j\in J^{\mathrm{miss}}$}
    \State \textbf{Step 2.1 (Design matrix construction):}
    \State $\bullet$ For each time index $t$, construct the predictor \State vector 
    \State $\boldsymbol{Z}_{t,-j}=
        \big(X_{t,-j},\;
        h_t,\;X_{t-1,j},\ldots,X_{t-\ell,j},\;
        X_{t+1,j},\ldots,X_{t+\ell,j}\big)$,
    \State using only available values from the current $\mathbf{X}$
     \State \textbf{Step 2.2 (Bayesian regression):}
    \State $\bullet$ Fit the Bayesian regression model 
      \State \[X_{t,j} = f_j(\boldsymbol{Z}_{t,-j}^\top \theta_j) + \varepsilon_{t,j},
        \quad \varepsilon_{t,j} \sim \mathcal{N}(0,\tau_j^2)\]
    \State  using observations $t \in I_j^{\mathrm{obs}}$ of $\mathbf{X}$.
     \State \textbf{Step 2.3 (Posterior sampling):}
     \State $\bullet$ Sample $(\theta_j, \tau_j^2)$ from the posterior
    \[
    p(\theta_j, \tau_j^2 \mid X_j^{\mathrm{obs}}, \boldsymbol{Z}_{-j})
    \]
    \State using an MCMC sampler (RWM or MALA), with 
    \State adaptive proposal scaling.
    
    \State \textbf{Step 2.4 (Imputation):}
    \State $\bullet$ For each $t \in I_j^{\mathrm{miss}}$, draw a predicted value $\widehat{x}_{t,j}$ from 
        \State the predictive distribution $\widehat{\mathbb{P}}_j(\boldsymbol{z}_{t,-j})$, where $\boldsymbol{z}_{t,-j}$ is
        \State  taken from $\mathbf{X}$, and impute it in place at position $(t,j)$ 
        \State of $\mathbf{X}$ using
         \[
        X_{t,j}^{\mathrm{mis}} \sim
        \mathcal{N}\!\left(
        \boldsymbol{Z}_{t,-j}^\top \theta_j,\;
        \tau_j^2
        \right),
        \]
        
        
    \EndFor
\EndFor  
\State Store $\mathbf{X}$
\State \textbf{Step 3 (Multiple imputations):}
\State Repeat \textbf{Step 1--2} with different random seeds to obtain $m$ completed datasets $\mathbf{X}_1,\ldots,\mathbf{X}_m$.
\State \textbf{Output:} Multiple imputed datasets for downstream analysis and pooling.
\end{algorithmic}    
\end{algorithm}

\begin{figure*}
    \centering
    \includegraphics[width=\linewidth]{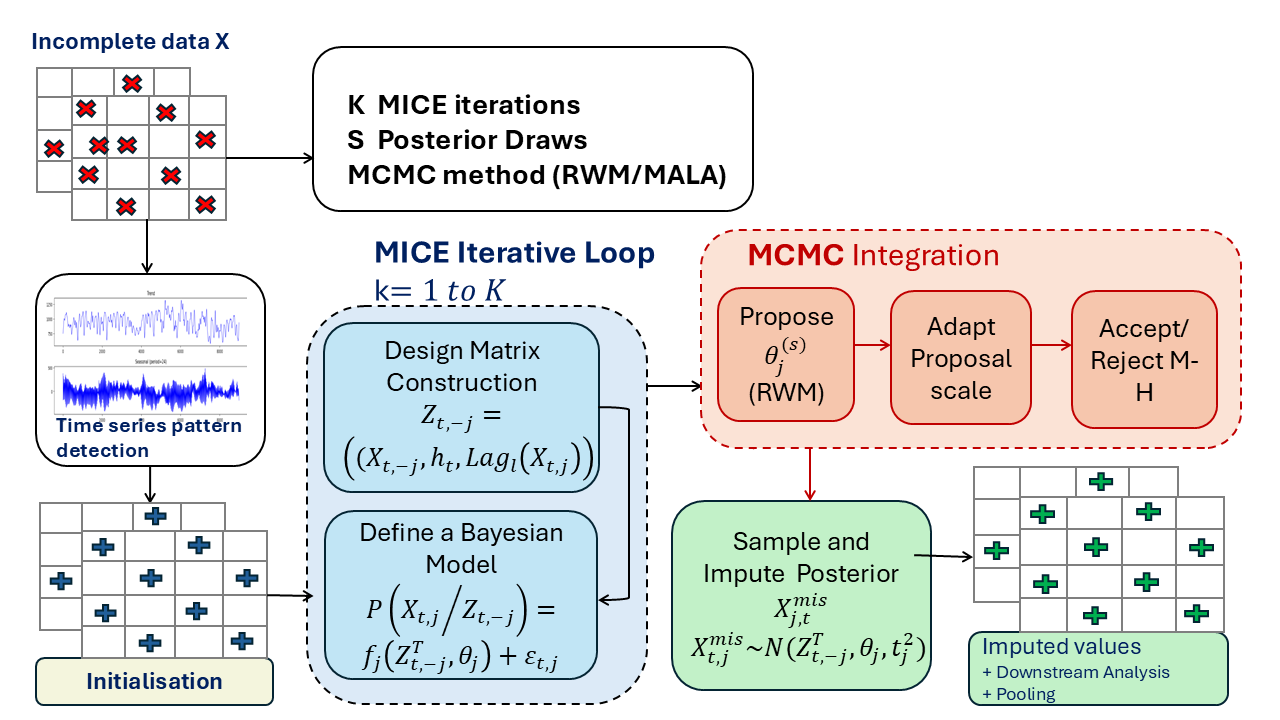}
    \caption{The proposed tBayes-MICE imputation framework. The framework consists of temporal pattern detection, placeholder initialisation, MICE loop with lagged predictors and Bayesian modelling, while MCMC (RWM or MALA) provides posterior sampling and parameter updates. The final imputation is generated via posterior predictive draws.}
    \label{fig:MCMC-MICE_Framework}
\end{figure*}

 Our tBayes-MICE framework, as shown in \cref{fig:MCMC-MICE_Framework} is accompanied by details given in Algorithm~2. The process begins with an incomplete time-series dataset $X$, in which missing observations are indicated by red crosses in \cref{fig:MCMC-MICE_Framework}. 

\paragraph{Step 1 (Initialisation)} A placeholder initialisation (time-aware or mean-based) temporarily fills in missing entries, providing a starting point for the iterative imputation. If the initialisation is time-aware, a preliminary time-series pattern-detection step captures temporal dependencies and autocorrelation structures to ensure that imputations preserve realistic dynamics.

\paragraph{Step 2 (Iterative imputation)} At each of the $K$ MICE iterations ($k = 1,..., K$), a target variable $X_{t,j}$ with missing values is selected. Its predictors consist of past and future lagged values of $X_{t,j}$ and contemporaneous variables $X_{t,-j}$ at current time $t$ formed as $Z_{t,-j}$, thereby accounting for temporal structure. A Bayesian model is then defined for $X_{t,j}$, specifying its conditional distribution given the predictors $Z_{t,-j}$. This formulation naturally incorporates prior uncertainty and supports probabilistic updating during imputation.
We sample (estimate) tBayes-MICE model parameters $\theta_j$ via MCMC  using either a Random Walk Metropolis (RWM) or a Metropolis-Adjusted Langevin Algorithm (MALA) scheme. Proposal scales are adaptively tuned, and acceptance is governed by the Metropolis-Hastings criterion, ensuring convergence to the posterior distribution. Once parameters are updated, the missing entries $X_{t,j}^{mis}$ are sampled from the posterior predictive distribution, $X_{t,j}^{mis} \sim N(Z_{t,-j}^{T} \theta_j, \tau_j^2)$. This guarantees that the imputations remain consistent with both the observed data and the estimated model parameters. 
\paragraph{Step 3 (Multiple imputation)} After completing all $K$ iterations and generating $S$ posterior draws, step 2 is repeated $m$ times with different random seeds to generate $m$ completed datasets. These are then pooled to produce the final imputed values suitable for downstream statistical analysis, modelling, and inference.

\subsection{Evaluation}

We assessed the imputation performance using two widely adopted error metrics: Normalised Root Mean Square (NRMSE) and Normalised Mean Absolute Error (NMAE). These metrics provide complementary insights into the accuracy and consistency of the imputation, measuring both absolute and relative deviations from the true values. Let $actual_i$ be the true (ground-truth) value of the $i$-th item, $pred_i$ the corresponding imputed (predicted) value, and $N$ the total number of imputed values. Then, NRMSE and NMAE are defined as follows:
\begin{equation}
    \mathrm{NRMSE} = \frac{\sqrt{\frac{1}{N}\sum_{i=1}^{N}(\mathrm{pred}_i - \mathrm{actual}_i)^2}}{\sigma_\text{actual}}
\end{equation}

\begin{equation}
  \mathrm{NMAE}
  = \frac{\frac{1}{N}\sum_{i=1}^{N}\left| \mathrm{pred}_i - \mathrm{actual}_i \right|}
     {\max(\mathrm{actual}_i)\;-\;\min(\mathrm{actual}_i)} 
\end{equation}


\subsection{Experimental Settings and MCMC Convergence Diagnostic}

 This section focuses on experimental settings and MCMC   convergence diagnostic criteria. We ran two independent chains per model (30{,}000 iterations each) and discarded the first 20\% as burn-in. We deliberately initialised the chains with over-dispersed starting values. For the parameter vector $\boldsymbol{\theta}$, we used proposal distribution with adaptive step sizes targeting acceptance rates of 0.20-0.30 (RWM) and 0.55-0.65 (MALA). We updated the residual variance $\tau^{2}$ via a Gibbs step in MALA and a Gaussian noise in RWM. We review the convergence visually (trace plots, autocorrelation functions, and running means) and with \texttt{Arviz} summaries: Gelman-Rubin $\hat{R}$, effective sample size (ESS; bulk), and standard deviation (SD). We declared convergence when $\hat{R} < 1.05$, $ESS_{\text{bulk}} > 400$ for key parameters, and $SD < 5\%$ of the posterior Standard deviation, consistent with common guidance from literature \cite{gelman1992inference}.

\section{Experimental Analysis and Results}

 We present the experimental evaluation and results, organised as follows:
\begin{itemize}
    \item We first compare the performance of univariate time-lagged MICE with established time-series baseline methods on a representative variable (WBC) from PhysioNet dataset.
    \item We then analyse the convergence behaviour of the two MCMC samplers to verify adequate posterior exploration and the reliability of the inferred results.
    \item Finally, we compare the proposed tBayes-MICE against MICE and BRITS across both datasets.
\end{itemize}

\subsection{Performance metrics of Univariate Time-Lagged MICE}
\textcolor{black}{We evaluated a univariate time-lagged MICE framework that conditions each missing observation on its immediate temporal neighbours ($\ell = 1$). The approach is compared against standard time-series imputation methods, including linear interpolation, $K$-Nearest Neighbours (KNN), mean and median imputation, seasonal decomposition, and Last Observation Carried Forward (LOCF).}

\begin{table}[!htbp]
\centering
\caption{Imputation performance comparison of univariate time-lagged MICE (lag = 1) with established time series baseline methods \textcolor{black}{ on a representative variable (WBC) from the PhysioNet dataset}. We report RMSE, MAE, NMAE, and NRMSE, Boldface highlights the best value among the methods.}
\label{tab:Univariate_MICE}
\small
\setlength{\tabcolsep}{6pt}
\renewcommand{\arraystretch}{1.15}
\begin{tabular}{lcccc}
\toprule
\textbf{Method} & \textbf{RMSE} & \textbf{MAE} & \textbf{NMAE} & \textbf{NRMSE} \\
\midrule
MICE (Time-Lagged) & \textbf{7.8500} & \textbf{5.0708} & \textbf{0.6005} & \textbf{0.9296} \\
Linear Interpolation           & 8.0579 & 5.2240 & 0.6186 & 0.9542 \\
KNN ($k{=}5$)                  & 8.4580 & 5.3885 & 0.6381 & 1.0016 \\
Mean Imputation                & 8.4580 & 5.3885 & 0.6381 & 1.0016 \\
Median Imputation              & 8.5415 & 5.1430 & 0.6090 & 1.0115 \\
Seasonal Decomposition         & 8.7597 & 5.2980 & 0.6274 & 1.0373 \\
LOCF                            & 10.6377 & 6.3747 & 0.7549 & 1.2597 \\
\bottomrule
\end{tabular}
\end{table}

\cref{tab:Univariate_MICE} summarises the imputation performance for the WBC feature of the dataset using the univariate time-lagged MICE model conditioned on immediate temporal neighbourhood ($\ell = 1$) alongside several classical time-series baseline methods. We observe that the univariate time-lagged MICE achieves the lowest error across all error metrics. In particular, it produced the smallest (best) RMSE (7.85) and MAE (5.07), while the normalised errors further support this observation with NRMSE (0.9296) and NMAE (0.6005), suggesting reliable performance relative to the data's variability. We further noticed that linear interpolation performed reasonably well compared to other baseline methods, while median yields the second-best result in terms of NMAE (0.6090). The poorest performance is observed for LOCF, which yields substantially larger errors across all metrics.

\subsection{MCMC Convergence Results}
We assess MCMC convergence using two independent chains per model, and the results are presented as follows.
\begin{figure*}[t]
  \centering
  \begin{subfigure}{0.49\textwidth}
    \includegraphics[width=\linewidth]{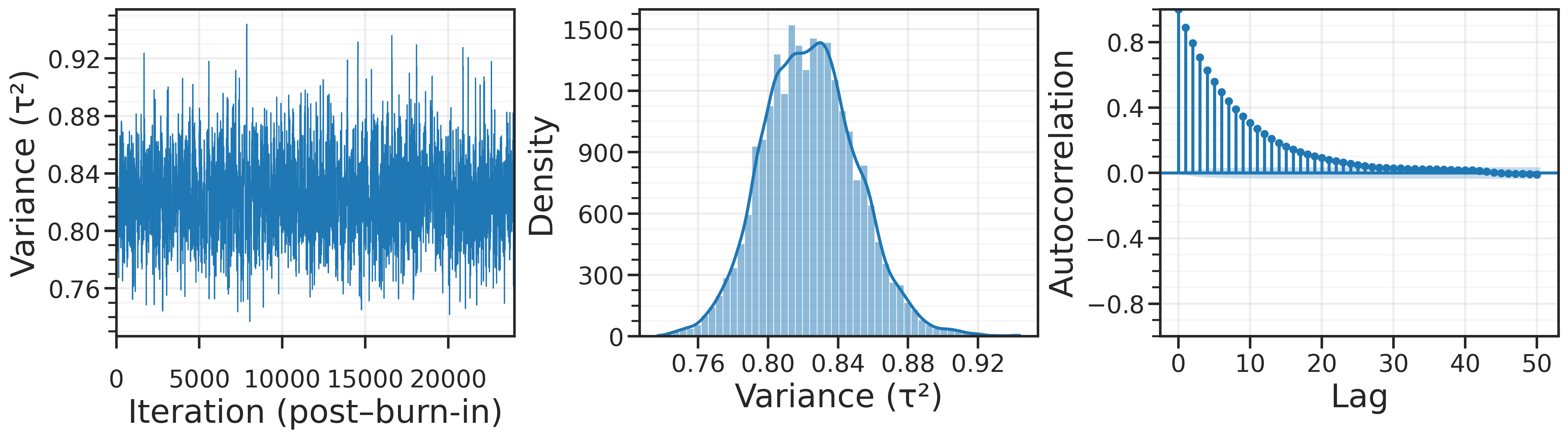}
    \caption{Chain 1 —  RWM MCMC}
  \end{subfigure}\hfill
  \begin{subfigure}{0.49\textwidth}
    \includegraphics[width=\linewidth]{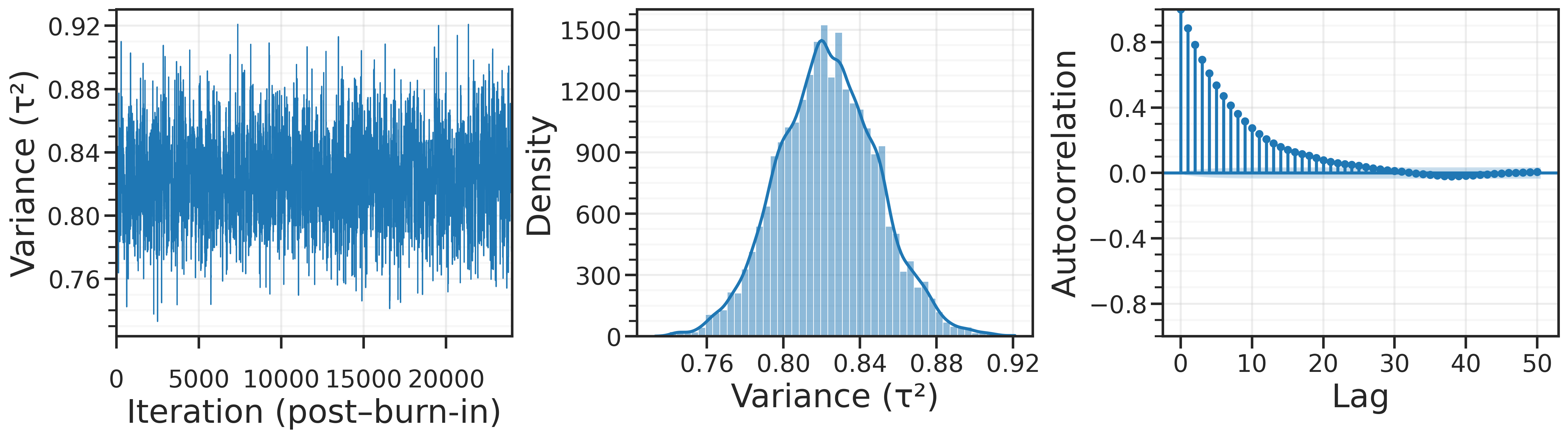}
    \caption{Chain 2 — RWM MCMC}
  \end{subfigure}

  \vspace{4pt}
  \begin{subfigure}{0.49\textwidth}
    \includegraphics[width=\linewidth]{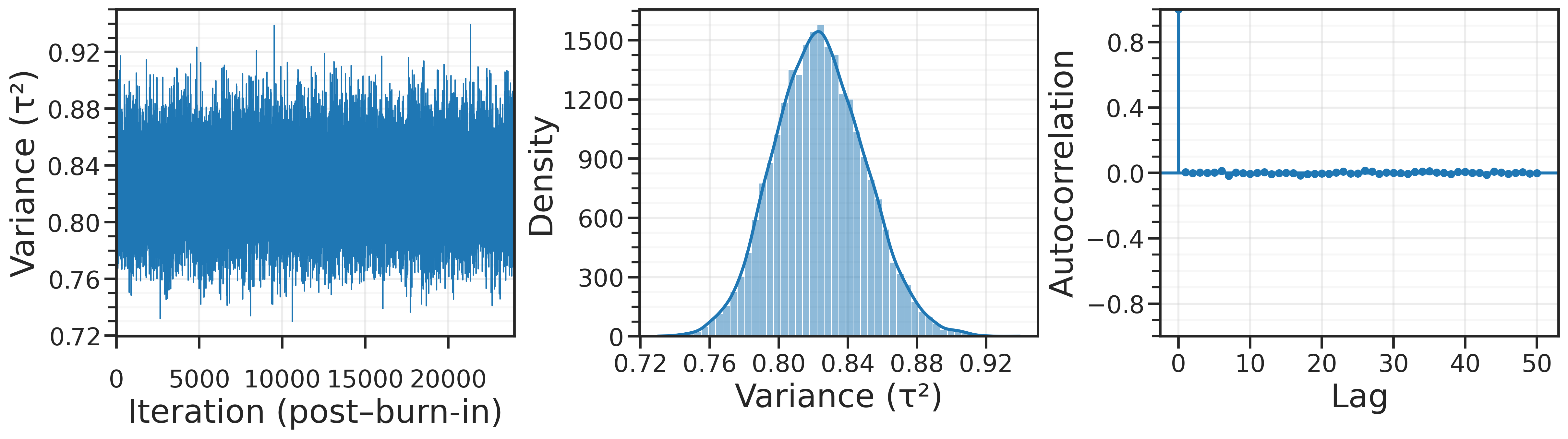}
    \caption{Chain 1 —  MALA MCMC}
  \end{subfigure}\hfill
  \begin{subfigure}{0.49\textwidth}
    \includegraphics[width=\linewidth]{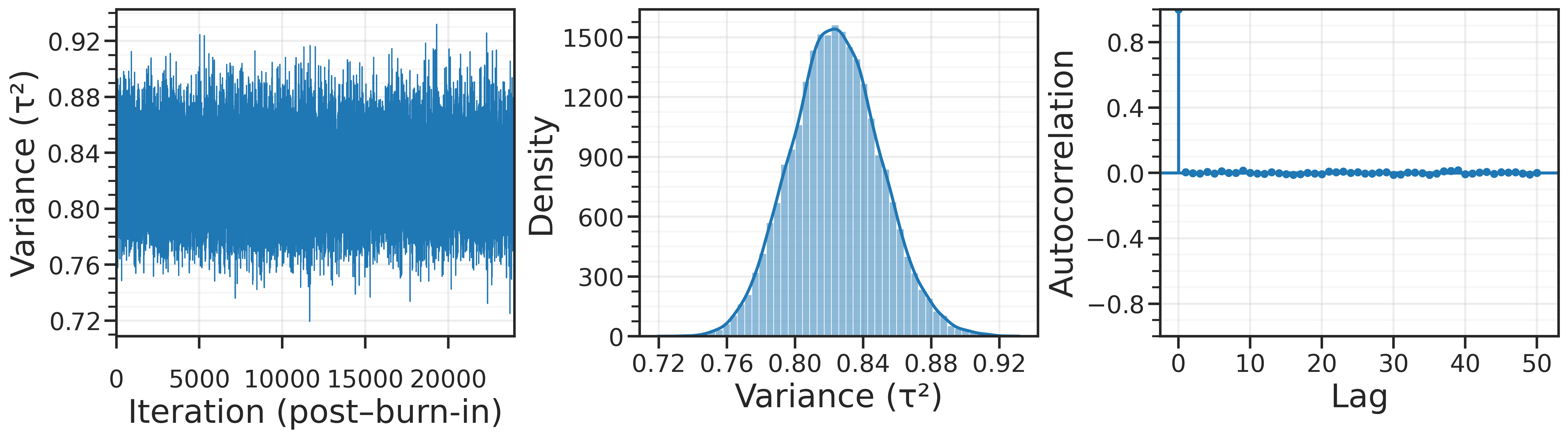}
    \caption{Chain 2 — MALA MCMC}
  \end{subfigure}

  \caption{Convergence diagnostics for $\tau^2$ (trace, marginal density, and ACF) across two chains and two samplers for one of the variables (\text{HC0\textsubscript{3}}) from the physionet dataset.}
  \label{fig:convergence_dual}
\end{figure*}

\cref{fig:convergence_dual} presents convergence diagnostics for the variance parameter $\tau^2$ obtained from two independent chains for each MCMC sampler: RWM (panels a-b) and MALA (panels c-d).  From these, it was evident that both sets of chains show stable trace behaviour post-burn-in and produce highly overlapping marginal posterior densities, thus demonstrating convergence to a common unimodal posterior distribution. Further, the posterior mass appears to be highly concentrated in a relatively narrow range, with both samplers producing very similar central tendency values; thus, both appear to be recovering consistent estimates of uncertainty for $\tau^2$. We notice, however, that the major difference lies in the mixing characteristics of the two samplers. It appears that the RWM samplers exhibit strong serial correlation (i.e., slowly decaying autocorrelation functions that remain significant at approximately 15-25 lags), resulting in smaller effective sample sizes than those of MALA samplers. By comparison, we observe that the MALA sampler exhibits excellent mixing, as indicated by its rapid decline in the autocorrelation function to nearly zero within a few lags. This resulted in "hairier" trace plots (as opposed to those of RWM) and indicates an improvement in the overall sampling efficiency of MALA over RWM while maintaining equivalent posterior inference for $\tau^2$.

\begin{table*}[t]
\centering
\caption{Convergence summaries for the bicarbonate (\text{HCO\textsubscript{3}}) variable from the PhysioNet dataset using tBayes-MICE. Results are shown for two MCMC samplers: MALA (left) and RWM (right). The table reports posterior means, standard deviations (SD), 94\% highest density intervals (HDIs), effective sample sizes (ESS), and $\hat{R}$ statistics.}
\begin{subtable}{0.49\textwidth}
\centering
\caption{MALA--MCMC}
\scriptsize
\setlength{\tabcolsep}{4pt}
\renewcommand{\arraystretch}{1.15}
\resizebox{\linewidth}{!}{%
\begin{tabular}{
l
S[table-format=1.3]
S[table-format=1.3]
S[table-format=2.3]
S[table-format=2.3]
S[table-format=5.0]
S[table-format=1.2]
}
\toprule
{Param} & {Mean} & {SD} & {HDI$_{3\%}$} & {HDI$_{97\%}$} & {ESS} & {$\hat{R}$} \\
\midrule
$b$    &  0.000 & 0.017 & -0.032 &  0.032 &  10609 & 1.00 \\
$\tau$ &  0.563 & 0.018 &  0.530 &  0.597 & 46291 & 1.00 \\
RMSE   &  0.75 & 0.001 &  0.748 &  0.752 &  8311 & 1.00 \\
\midrule
$w_1$  &  0.024 & 0.017 & -0.007 &  0.056 &  9622 & 1.00 \\
$w_2$  &  0.073 & 0.018 &  0.039 &  0.106 &  7709 & 1.00 \\
$w_3$  &  0.082 & 0.018 &  0.047 &  0.117 &  6663 & 1.00 \\
$w_4$  & -0.094 & 0.018 & -0.128 & -0.058 &  6812 & 1.00 \\
$w_5$  &  0.049 & 0.017 &  0.017 &  0.081 &  9647 & 1.00 \\
$w_6$  &  0.004 & 0.017 & -0.027 &  0.035 &  10891& 1.00 \\
\bottomrule
\end{tabular}}
\end{subtable}
\hfill
\begin{subtable}{0.49\textwidth}
\centering
\caption{RWM--MCMC}
\scriptsize
\setlength{\tabcolsep}{4pt}
\renewcommand{\arraystretch}{1.15}
\resizebox{\linewidth}{!}{%
\begin{tabular}{
l
S[table-format=1.3]
S[table-format=1.3]
S[table-format=2.3]
S[table-format=2.3]
S[table-format=5.0]
S[table-format=1.2]
}
\toprule
{Param} & {Mean} & {SD} & {HDI$_{3\%}$} & {HDI$_{97\%}$} & {ESS} & {$\hat{R}$} \\
\midrule
$b$    & 0.000 & 0.017 & -0.030 &  0.033 &  1070 & 1.00 \\
$\tau$ &  0.564 & 0.018 &  0.531 &  0.598 & 2539 & 1.00 \\
RMSE   &  0.751 & 0.001 &  0.749 &  0.753 &  870 & 1.00 \\
\midrule
$w_1$  &  0.025 & 0.017 & -0.006 & 0.056&  956 & 1.00 \\
$w_2$  &  0.071 & 0.018 & 0.040 &  0.106 &  754 & 1.00 \\
$w_3$  &  0.080 & 0.019 & 0.044 & 0.116 &  601 & 1.00 \\
$w_4$  & -0.093 & 0.019 & -0.128 & -0.057 & 631 & 1.00 \\
$w_5$  &  0.051 & 0.018 & 0.018 & 0.083 &  900 & 1.00 \\
$w_6$  &  0.004 & 0.018 & -0.029 & 0.037 &  1070 & 1.00 \\
\bottomrule
\end{tabular}}
\end{subtable}
\label{tab:convergence_summary}
\end{table*}

We notice that in addition to the similar results from the visual evidence (see \cref{fig:convergence_dual}), the convergence diagnostics summarise in \cref{tab:convergence_summary} also show that both RWM and MALA samplers have nearly identical posterior means and uncertainty intervals for all of the reported parameters, which indicates that they converge to the same stable posterior distribution. However, we find that the MALA sampler was significantly more efficient than the RWM sampler, as evidenced by the fact that MALA achieves a substantially higher ESS\textsubscript{bulk} (values greater than 5,000 for almost every parameter; values greater than 46,000 for the variance parameter ($\tau^2$)). The high ESS values reflect low autocorrelation and efficient exploration of the posterior space. In contrast, the RWM sampler produces lower ESS values (typically less than 1,000) for almost all parameters, indicating stronger serial dependence and slower mixing. It is clear that this is true, especially when examining the regression coefficients, as the MALA sampler produces an order-of-magnitude increase in effective samples than the RWM. Furthermore, regardless of the efficiency differences observed between the two samplers, all $\hat{R}$ statistics were equal to 1.00 for both samplers, thereby demonstrating good convergence across all chains. Thus, the table clearly illustrates that while the two samplers produce the same posterior structure, the MALA sampler is substantially more efficient and therefore able to produce more independent information per iteration, thereby converging faster given the same computational budget.

\subsection{Performance metrics of the models}
We compared the performance of the proposed tBayes-MICE with the standard MICE and BRITS baseline models, and present the results below

\begin{table*}[!htbp]
\centering
\caption{Imputation performance of MICE, BRITS and tBayes-MICE under two MCMC samplers (RWM/MALA). Metrics are NMAE and NRMSE (mean, standard deviation) across 30 experimental runs for six AirQuality variables. Boldface highlights the best value per row.}
\small
\setlength{\tabcolsep}{6pt}
\renewcommand{\arraystretch}{1.15}
\begin{tabular}{llcccc}
\toprule
 &  & \multicolumn{2}{c}{\textbf{RWM}} & \multicolumn{2}{c}{\textbf{MALA}} \\
\cmidrule(lr){3-4}\cmidrule(lr){5-6}
\textbf{Variable} & \textbf{Method}
& \textbf{NMAE} & \textbf{NRMSE}
& \textbf{NMAE} & \textbf{NRMSE} \\
\midrule
\multirow{4}{*}{CO(GT)}
 & MICE           & 0.2622 (0.0071) & 0.3376 (0.0083) & 0.2622 (0.0071) & 0.3376 (0.0083) \\
 & BRITS          & 0.2646 (0.0473) & 0.3870 (0.0653) & 0.2646 (0.0473) & 0.3870 (0.0653) \\
 & tBayes-MICE & \textbf{0.1349 (0.0013)} & \textbf{0.1788 ( 0.0017)}
                   & \textbf{0.1346 (0.0011)} & \textbf{0.1788 (0.0015)} \\
\midrule
\multirow{4}{*}{PT08.S1(CO)}
 & MICE           & 0.3811 (0.0097) & 0.4881 (0.0125) & 0.3811 (0.0097) & 0.4881 (0.0125) \\
 & BRITS          & 0.3418 (0.0240) & 0.6153 (0.0252) & 0.3418 (0.0240) & 0.6153 (0.0252) \\

 & tBayes-MICE &  \textbf{0.1788 (0.0018)} & \textbf{0.2648 (0.0019)}
                   & \textbf{0.1787 (0.0015)} & \textbf{0.2651 (0.0022)} \\
\midrule
\multirow{4}{*}{NMHC(GT)}
 & MICE           & 0.4778 (0.0131) & 0.6220 (0.0157)  & 0.4778 (0.0131) & 0.6220 (0.0157) \\
 & BRITS          & 0.3377 (0.0194)& 0.5586 (0.0353) & 0.3377 (0.0194)& 0.5586 (0.0353) \\

 & tBayes-MICE & \textbf{0.2201 (0.0021)} & \textbf{0.3449 (0.0030)}
                   & \textbf{0.2196 (0.0018)} & \textbf{0.3444 (0.0019)} \\
\midrule
\multirow{4}{*}{C\textsubscript{6}H\textsubscript{6}(GT)}
 & MICE           & 0.1687 (0.0041) & 0.2306 (0.0060) & 0.1687 (0.0041) & 0.2306 (0.0060) \\
 & BRITS          & 0.2855 (0.0545) & 0.4621 (0.0648) & 0.2855 (0.0545) & 0.4621 (0.0648) \\

 & tBayes-MICE & \textbf{0.1053 (0.0005)} & \textbf{0.1560 (0.0011)}
                   & \textbf{0.1055 (0.0007)} & \textbf{0.1562 (0.0014)} \\
\midrule
\multirow{4}{*}{PT08.S2(NMHC)}
 & MICE           & 0.2104 (0.0045) & 0.2811 (0.0060) & 0.2104 (0.0045) & 0.2811 (0.0060) \\
 & BRITS          & 0.2723 (0.0390) & 0.4541 (0.0380) & 0.2723 (0.0390) & 0.4541 (0.0380) \\

 & tBayes-MICE & \textbf{0.1183 (0.0005)} & \textbf{0.1671 (0.0012)}
                       & \textbf{0.1182 (0.0005)} & \textbf{0.1672 (0.0012)} \\
\midrule
\multirow{4}{*}{T}
 & MICE           & 0.9968 (0.0228)& 1.2539 (0.0285) & 0.9968 (0.0228)& 1.2539 (0.0285) \\
 & BRITS          & 0.3144 (0.0190) & 0.5653 (0.0194) & 0.3144 (0.0190) & 0.5653 (0.0194) \\
 
 & tBayes-MICE & \textbf{0.1383 (0.0013)} & \textbf{0.1932 (0.0014)}
                   & \textbf{0.1378 (0.0008)} & \textbf{0.1925 (0.0010)} \\
\bottomrule
\end{tabular}
\label{tab:metrics_rwm_mala_air}
\end{table*}

\begin{table*}[!htbp]
\centering
\caption{Imputation performance of MICE, BRITS and tBayes-MICE under two MCMC samplers (RWM/MALA). Metrics are NMAE and NRMSE (mean, standard deviation) across 30 experimental runs for four PhysioNet variables. Boldface highlights the best value per row.}
\small
\setlength{\tabcolsep}{6pt}
\renewcommand{\arraystretch}{1.15}
\begin{tabular}{llcccc}
\toprule
 &  & \multicolumn{2}{c}{\textbf{RWM}} & \multicolumn{2}{c}{\textbf{MALA}} \\
\cmidrule(lr){3-4}\cmidrule(lr){5-6}
\textbf{Variable} & \textbf{Method} 
& \textbf{NMAE} & \textbf{NRMSE}
& \textbf{NMAE} & \textbf{NRMSE} \\
\midrule
\multirow{4}{*}{HCO3}
 & MICE           & 1.1904 (0.0110) & 1.4997 (0.0128) & 1.1904 (0.0110) & 1.4997 (0.0128) \\
 & BRITS          & 0.7535 (0.0039) & 0.9941 (0.0049) & 0.7535 (0.0039) & 0.9941 (0.0049) \\
 & tBayes-MICE & \textbf{0.7189 (0.0018)} & \textbf{0.9375 (0.0021)}
                   & \textbf{0.7198 (0.0019)} & \textbf{0.9380 (0.0022)} \\
\midrule
\multirow{4}{*}{Na}
 & MICE           & 1.1467 (0.0098) & 1.4549 (0.0122) & 1.1467 (0.0098) & 1.4549 (0.0122) \\
 & BRITS          & 0.7732 (0.0023) & 1.5002 (0.0018) & 0.7732 (0.0023) & 1.5002 (0.0018) \\
 & tBayes-MICE & \textbf{0.7724 (0.0024)} & \textbf{1.0246 (0.0032)}
                   & \textbf{0.7715 (0.0018)} & \textbf{1.0231 (0.0029)} \\
\midrule
\multirow{4}{*}{Platelet}
 & MICE           & 1.1457 (0.0083) & 1.4816 (0.0093) & 1.1457 (0.0083) & 1.4816 (0.0093) \\
 & BRITS          & \textbf{0.6561 (0.0035)} & 0.9534 (0.0045) & \textbf{0.6561 (0.0035)} & 0.9534 (0.0045) \\
 & tBayes-MICE & 0.6596 (0.0019) & \textbf{0.9465 (0.0018)}
                   & 0.6601 (0.0022) & \textbf{0.9470 (0.0019)} \\
\midrule
\multirow{4}{*}{WBC}
 & MICE           & 1.2029 (0.0080) & 1.6035 (0.0125) & 1.2029 (0.0080) & 1.6035 (0.0125) \\
 & BRITS          & \textbf{0.5758 (0.0040)} &  0.9658 (0.0057) &
                     \textbf{0.5758 (0.0040)} &  0.9658 (0.0057)  \\
 & tBayes-MICE & 0.6228 (0.0022) & \textbf{0.9039 (0.0026)}
                   & 0.6224 (0.0030) & \textbf{0.9040 (0.0028)} \\
\bottomrule
\end{tabular}
\label{tab:metrics_rwm_mala_physio}
\end{table*}

\cref{tab:metrics_rwm_mala_air} presents the imputation performance of MICE, BRITS, and the proposed tBayes-MICE across six AirQuality variables under both RWM and MALA samplers. The results indicate that tBayes-MICE provide the smallest values for both NMAE and NRMSE across all variables compared to the other two methods.  This shows that tBayes-MICE substantially outperforms both MICE and BRITS across all variables. Specifically, there was a substantial reduction in both absolute and squared errors, which indicates greater accuracy and better control over large deviations during the imputation process. We further note that the improvements from tBayes-MICE were consistent across RWM and MALA samplers, with only marginal differences between them. The consistency of the improvement across both samplers indicates that the primary source of benefit lies in the Bayesian modelling framework used and the uncertainty-aware imputation strategy employed in tBayes-MICE, rather than the sampler. Conversely, the classical MICE produce the largest NMAE and NRMSE across all variables evaluated. Although BRITS exhibited significant improvements relative to the classical version of MICE, it remains inferior to the tBayes-MICE. Finally, we also observe that tBayes-MICE exhibit significantly lower standard deviations across the 30 experimental runs, demonstrating improved stability and robustness of this algorithm.

\cref{tab:metrics_rwm_mala_physio} summarises the imputation performance on the PhysioNet dataset under two MCMC samplers, comparing deterministic baselines (MICE and BRITS) with the proposed tBayes-MICE model. We report the mean and standard deviation of NMAE and NRMSE across 30 experimental runs.  Compared with the deterministic methods (MICE and BRITS), we observe that tBayes-MICE performs best across most of the clinical variables, regardless of the sampling scheme. While we see that MICE results in the largest errors across all clinical variables, this clearly illustrates that MICE has limitations in capturing temporal information and propagating its uncertainty within clinical time series data. In contrast, we find that BRITS significantly improves upon MICE across all clinical variables; therefore, highlighting the benefit of recurrent modelling for time-dependent physiological signals. Moreover, we find that tBayes-MICE yields the smallest NMAE and NRMSE for HCO\textsubscript{3} and Na, and competitive NRMSE for platelet and WBC. Conversely, for platelets and WBC, BRITS attains the smallest NMAE, and tBayes-MICE achieves the lowest NRMSE, indicating better overall imputation error control. Finally, regarding the samplers, we note that MALA produces nearly identical posterior estimates to RWM as in the AirQuality dataset; however, we do observe a slight difference between the samplers. This behaviour is consistent with the convergence diagnostic and indicates that both samplers recovered a very similar posterior structure.

\begin{figure*}[!htbp]
  \centering
  \begin{subfigure}{0.9\textwidth}
    \includegraphics[width=0.9\linewidth]{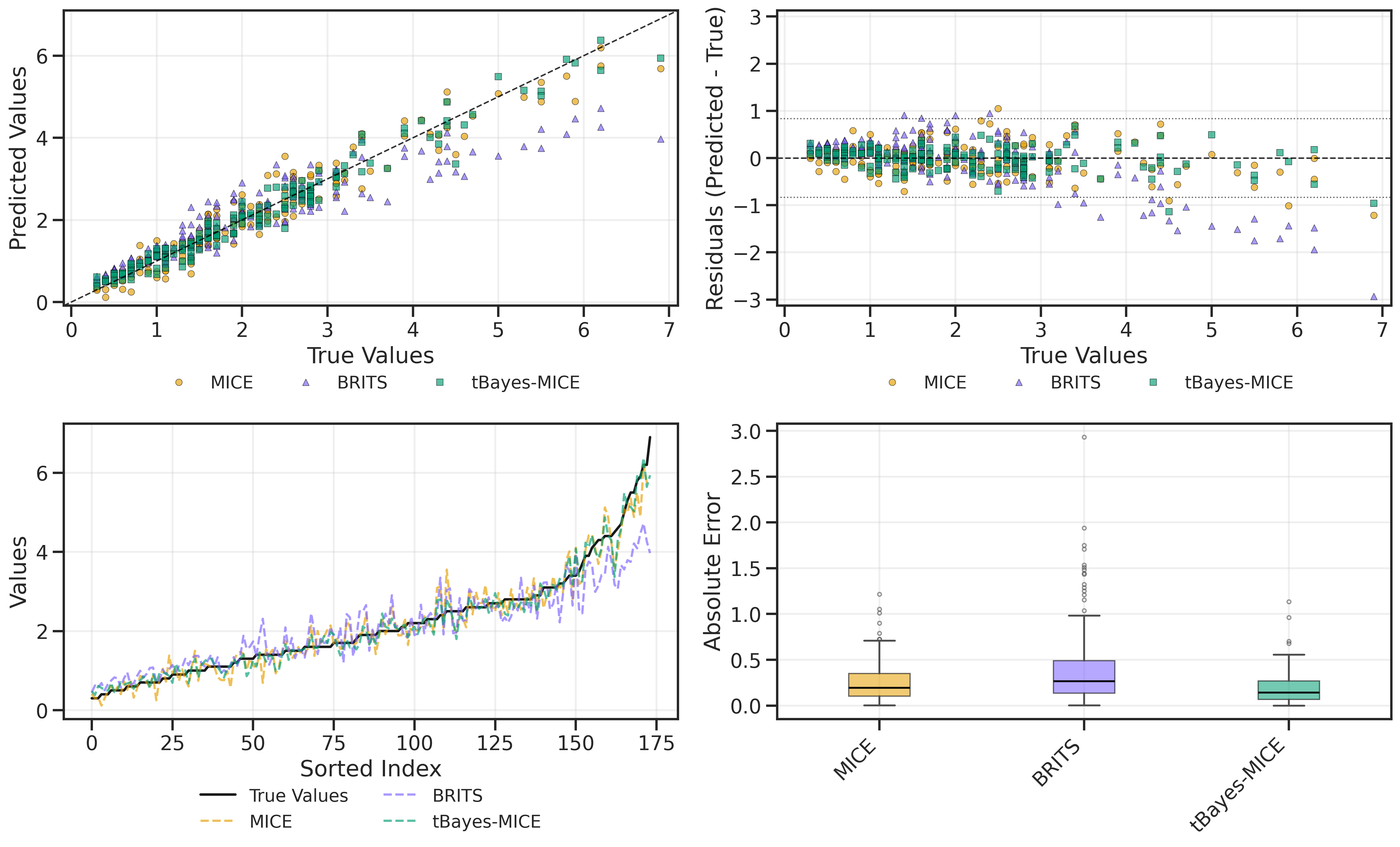}
   \caption{Prediction accuracy comparison for CO(GT) from AirQuality dataset.}
  \end{subfigure}

  \vspace{4pt}
  \begin{subfigure}{0.9\textwidth}
    \includegraphics[width=0.9\linewidth]{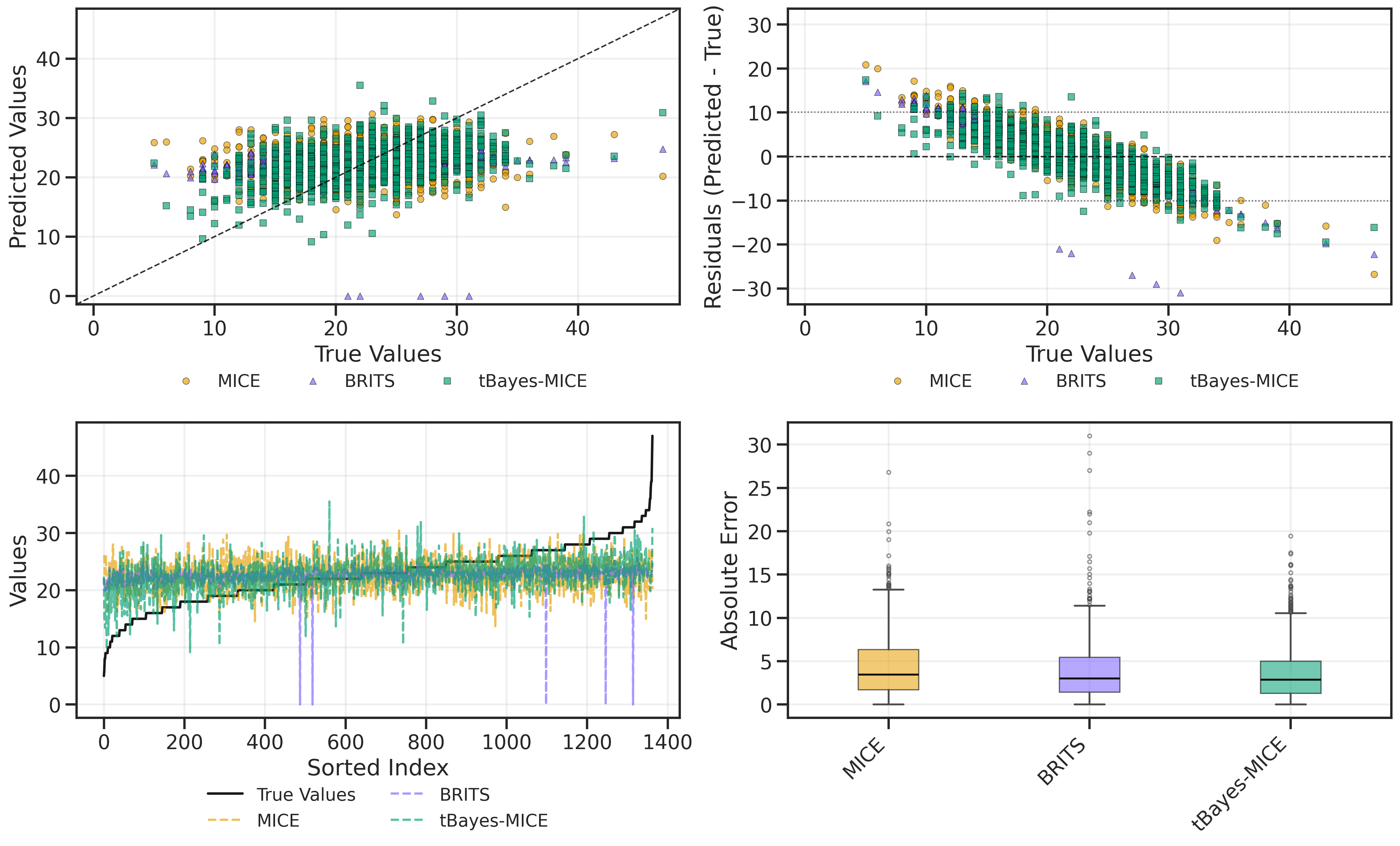}
   \caption{Prediction accuracy comparison for HCO\textsubscript{3} from physionet dataset.} 
  \end{subfigure}

  \caption{Prediction accuracy comparison for the CO(GT) variable from the AirQuality dataset (top) and the HCO\textsubscript{3} variable from the physioNet dataset (bottom). Each subplot panel shows:(top-left) predicted versus true values, (top-right) residuals against the true values, (bottom-left) sorted true values with corresponding model predictions, and (bottom-right) box plots of absolute errors.}
  \label{fig:prediction_accuracy}
\end{figure*}

\cref{fig:prediction_accuracy} compares prediction accuracy across imputation methods for two representative Variables: CO(GT) from the AirQuality dataset (plot a) and HCO\textsubscript{3} from the PhysioNet dataset (plot b). Each figure presents four complementary views: predicted versus true values, residuals against true values, sorted true values with corresponding model predictions, and box plots of absolute errors. For the AirQuality CO(GT) variable, we notice that the tBayes-MICE predictions closely followed the reference line, indicating strong agreement with the true values across the full range of the data. On the other hand, the Standard MICE and BRITS methods show much greater dispersion, especially at higher concentrations. The residual plots also indicated that the tBayes-MICE predictions were much more tightly centred around zero, whereas the deterministic baseline models show much greater variance. Additionally, this behaviour could further be seen in the sorted-value plots. The tBayes-MICE predictions track the true signal much more consistently, while MICE and BRITS show noticeable deviations from the true signal in several regions. The box plots of the absolute error support this conclusion, showing that tBayes-MICE yields a lower median absolute error and less variability compared to the baselines.

For the PhysioNet HCO\textsubscript{3} variable, we observe a similar type of behaviour, but with a much larger dynamic range because of the greater overall variability of the data. Standard MICE shows an increasing bias as the true values increase, whereas BRITS produces several large deviations from the true signal trajectory. In contrast, tBayes-MICE predictions remain closer to the true values across the entire sorted index, capturing the signal's global trends and local fluctuations more effectively. Furthermore, the residual plots of the tBayes-MICE indicate a decrease in the number of extreme negative errors, lower median absolute errors, and tighter interquartile ranges compared to MICE and BRITS.

\begin{figure*}[!htbp]
  \centering
  \begin{subfigure}{0.9\textwidth}
    \includegraphics[width=\linewidth]{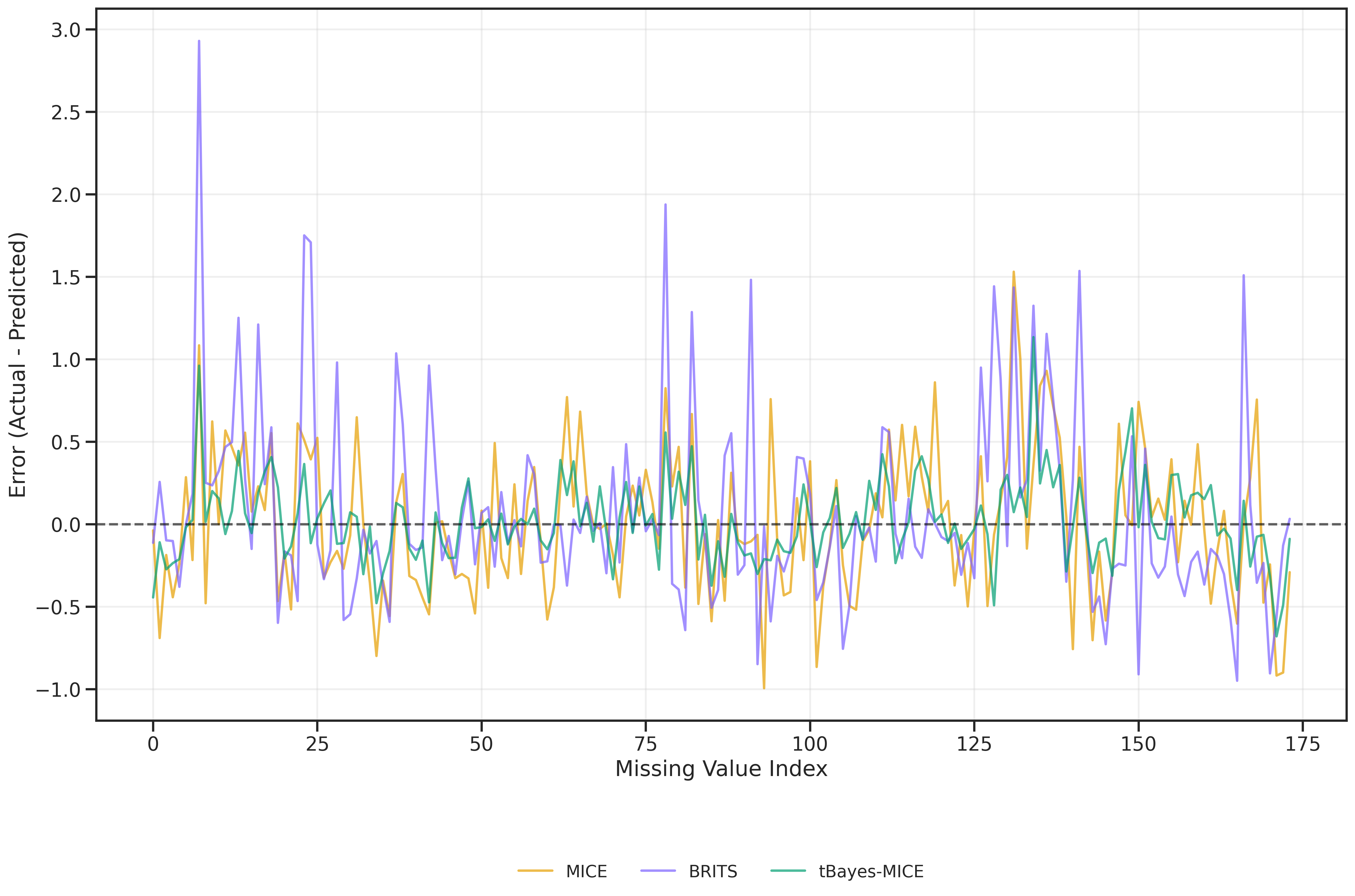}
   \caption{Imputation error across all methods for CO(GT) from AirQuality dataset}
  \end{subfigure}

  \vspace{4pt}
  \begin{subfigure}{0.9\textwidth}
    \includegraphics[width=\linewidth]{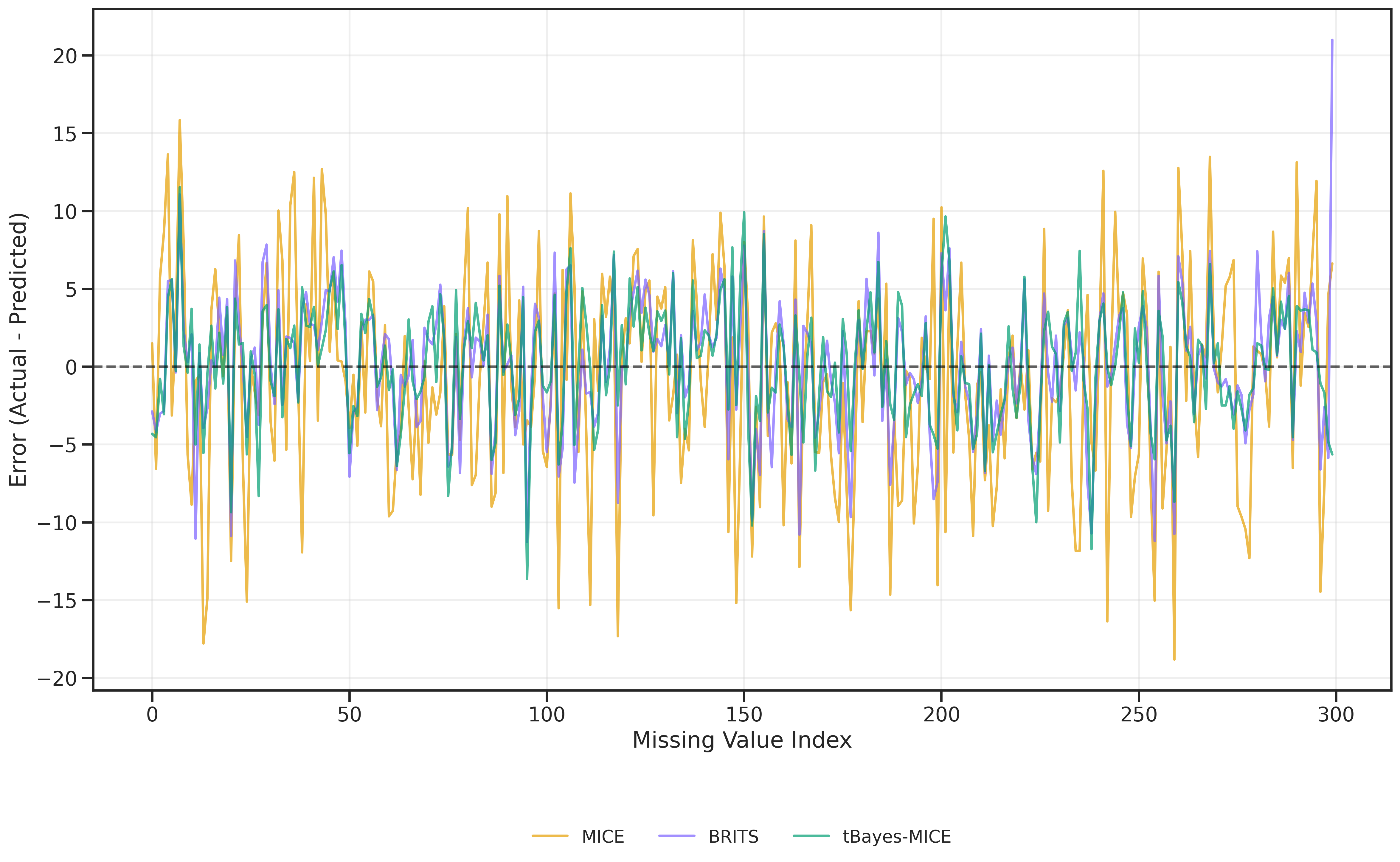}
   \caption{Imputation error across all methods for HCO\textsubscript{3} from physionet dataset} 
  \end{subfigure}

  \caption{Imputation error patterns over time index of each method across different datasets, with the AirQuality CO(GT) variable shown in the top panel and the PhysioNet HCO\textsubscript{3} variable shown in the bottom panel}
  \label{fig:imputation_error}
\end{figure*}

\cref{fig:imputation_error} illustrates the imputation error pattern over time across methods for two representative variables: CO(GT) from the AirQuality dataset (Fig.~5a) and HCO\textsubscript{3} from the PhysioNet dataset (Fig. 5b). The figure reveals clear discrepancies in both magnitude and dispersion across methods and datasets. For the AirQuality dataset (Fig. 5a), we observe that tBayes-MICE produce error trajectories that are tightly concentrated around zero with relatively small fluctuations across missing indices. On the other hand, standard MICE shows greater variation, while BRITS exhibit more pronounced error spikes and greater dispersion. For the PhysioNet HCO\textsubscript{3} variable (Fig.~5b), we observe a similar pattern, though with larger error magnitudes due to higher intrinsic variability of the data. The tBayes-MICE again produce more concentrated error distributions with fewer extreme deviations, whereas BRITS exhibits heavier tails, and standard MICE display a wider spread of errors despite being centred around zero on average. Across both datasets, we find that tBayes-MICE consistently reduces both error magnitude and variability, resulting in smoother and more stable error profiles.

\begin{figure*}[t]
  \centering
  \begin{subfigure}{0.49\textwidth}
    \includegraphics[width=\linewidth]{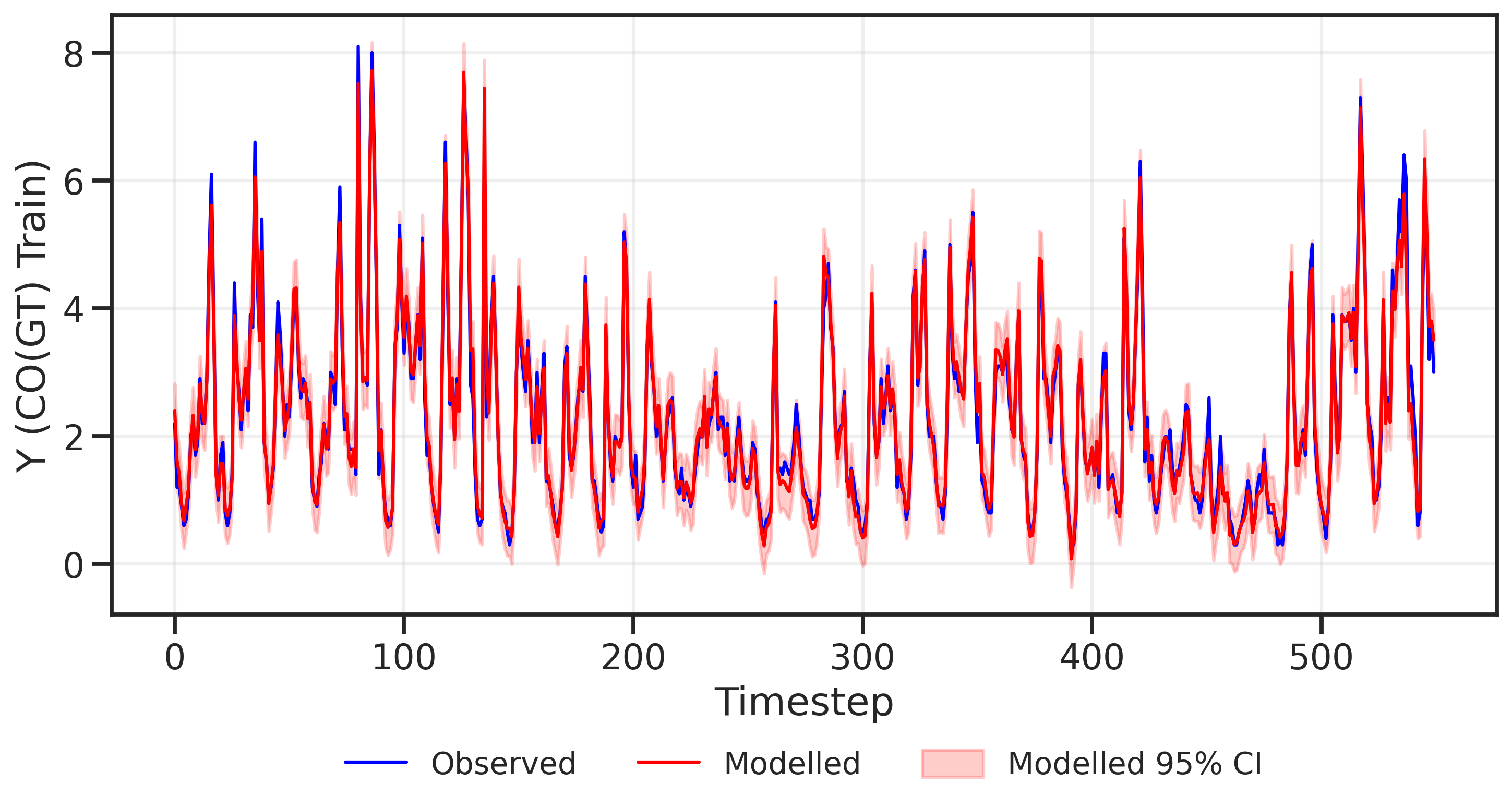}
    \caption{Train CO(GT) variable}
  \end{subfigure}\hfill
  \begin{subfigure}{0.49\textwidth}
    \includegraphics[width=\linewidth]{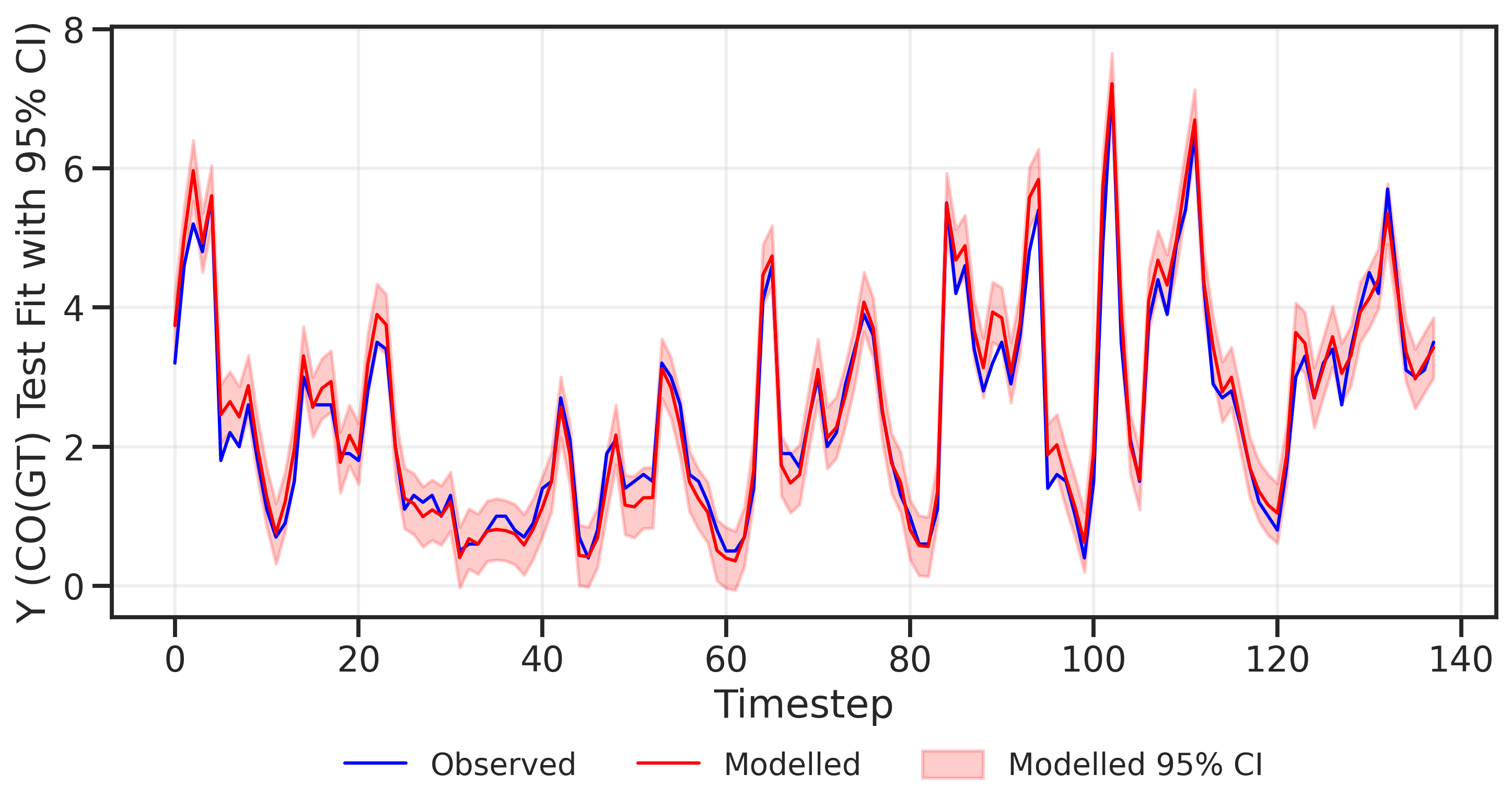}
    \caption{Test CO(GT) variable}
  \end{subfigure}

  \vspace{4pt}
  \begin{subfigure}{0.49\textwidth}
    \includegraphics[width=\linewidth]{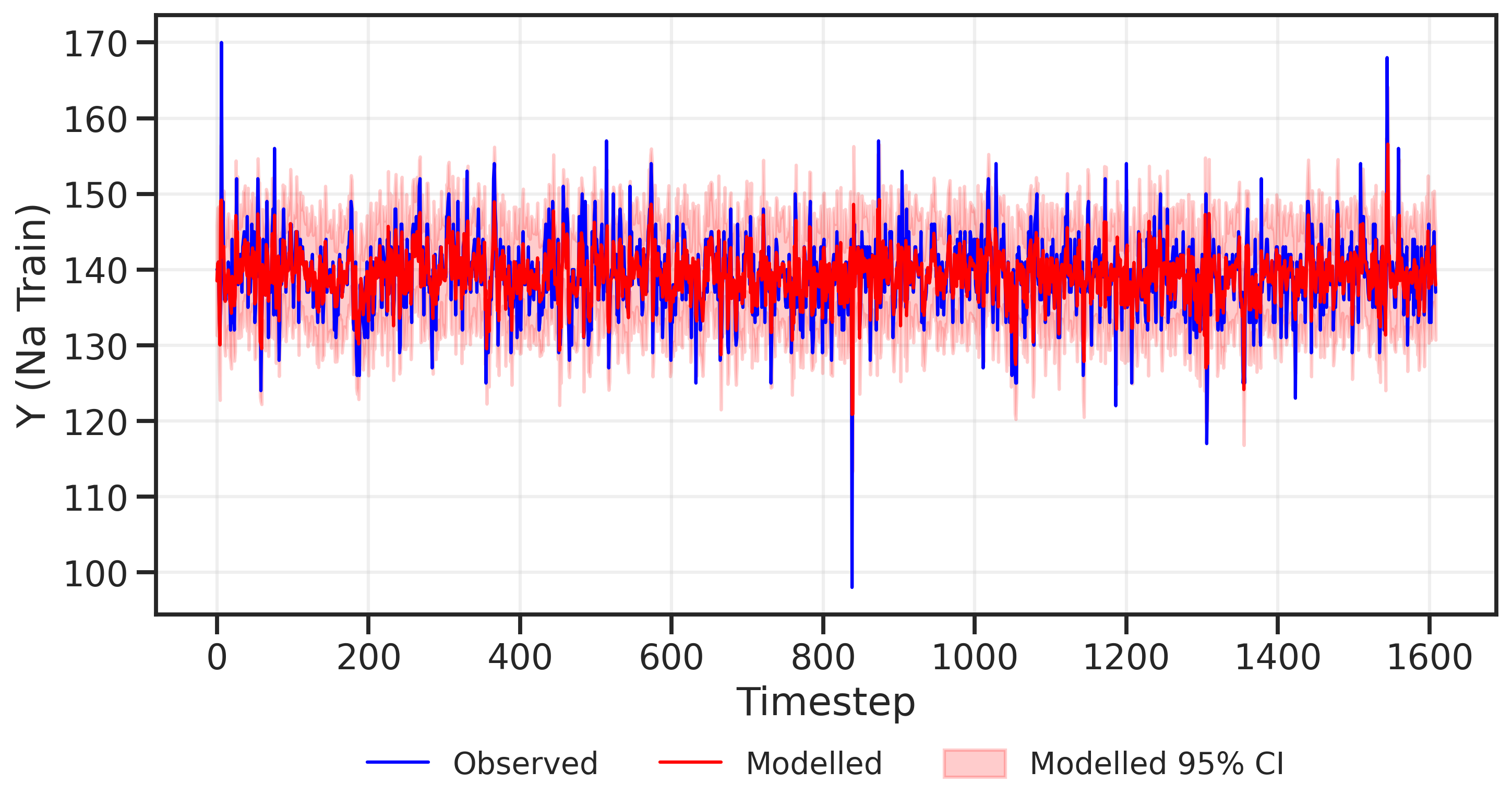}
    \caption{Train Na variable}
  \end{subfigure}\hfill
  \begin{subfigure}{0.49\textwidth}
    \includegraphics[width=\linewidth]{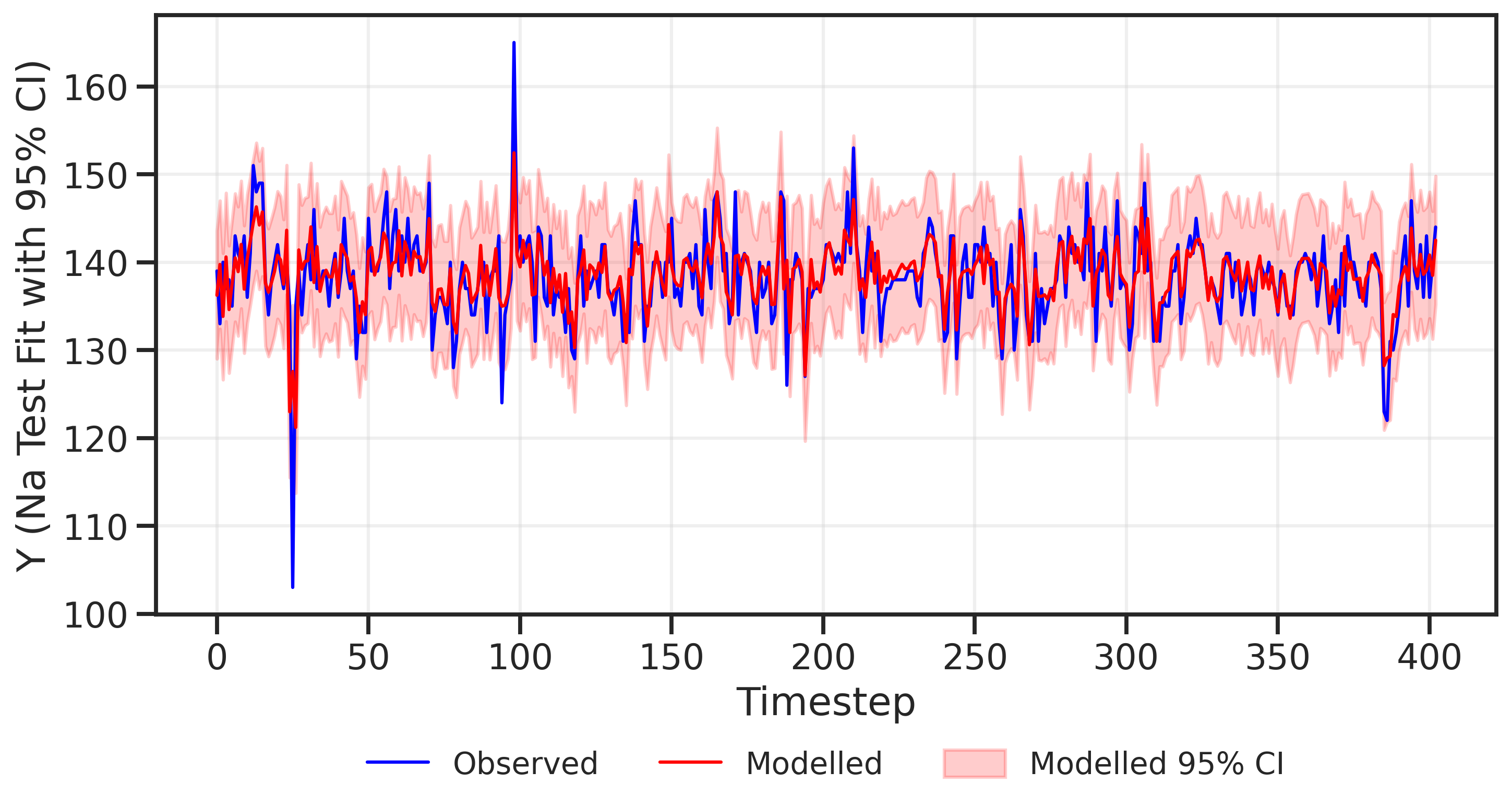}
    \caption{Test Na variable}
  \end{subfigure}

  \caption{Posterior predictive performance of the proposed tBayes-MICE model over time for the environmental CO(GT) variable (top row) and the clinical Na variable (bottom row), shown across training(left) and test(right) sets. The blue lines represent observed values, the red lines indicate posterior predictive means, and the shaded regions denote the 95\% credible intervals.}
  \label{fig:train_test}
\end{figure*}
\cref{fig:train_test} presents the posterior predictive performance of our proposed tBayes-MICE model for two representative variables, CO(GT) (environmental) and Na (clinical), across both training and test sets. As shown in the Figure, the model closely follows the observed CO(GT) temporal trajectory, accurately capturing the rapid fluctuations and seasonal-like spikes typical of air-quality dynamics. We further notice that the 95\% credible interval bands remain relatively tight during the training phase and widen slightly in the test phase, reflecting the expected increase in uncertainty when predicting unseen data. Despite this widening, the predictive mean remains well aligned with the true CO(GT) signals, accurately tracking both sharp rises and abrupt drops, which indicates strong generalisation and model stability.
A similar behaviour is observed in the Na values as the model closely tracks the observed values across both the training and test intervals. We also notice that the credible intervals are wider than those for CO(GT), indicating that the model appropriately captures the higher intrinsic variability and longer temporal dependence of the clinical data. However, even with this increased uncertainty, the predictive mean consistently remains centred on the true Na trajectory, confirming robust learning and reliable generalisation in the presence of physiological noise and temporal variability.

\begin{figure*}[!htbp]
  \centering
    \includegraphics[width=\linewidth]{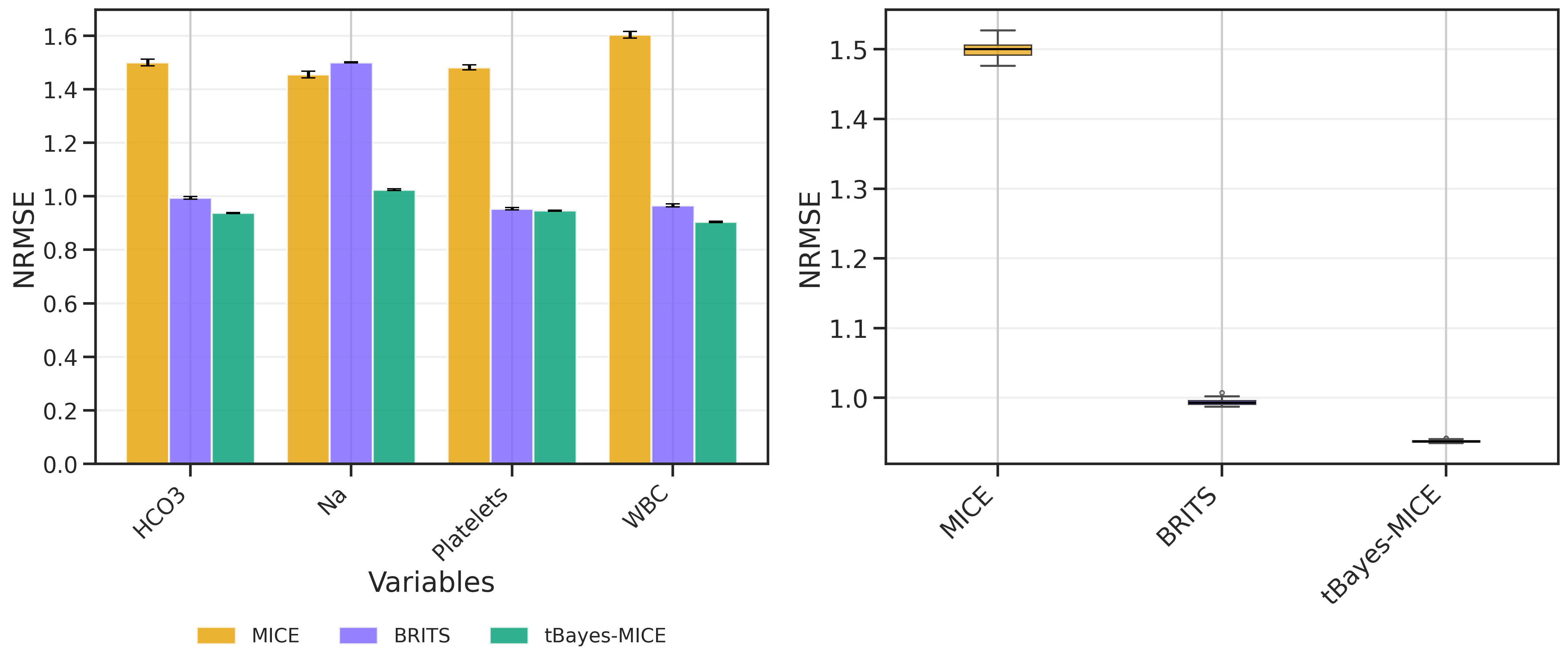}
  \caption{Comparative imputation accuracy and distribution across runs (box plot) for deterministic and Bayesian methods on the PhsyioNet dataset using NRMSE}
  \label{fig:experiment_summary}
\end{figure*}

\cref{fig:experiment_summary} illustrates the comparative imputation accuracy and distribution of results across 30 experimental runs for the methods on the PhysioNet dataset. The figure shows that tBayes-MICE consistently achieves substantially lower NRMSE than standard MICE and BRITS across all variables, with particularly pronounced improvements for HCO\textsubscript{3}, Na, and WBC. From the aggregated boxplot, we notice that tBayes-MICE not only reduce NRMSE but also exhibit markedly lower variability across runs compared to deterministic methods. Although BRITS improves upon standard MICE for some variables, the Bayesian approaches still outperformed them in both accuracy and stability.

\subsection{Ablation Study}
We conduct an ablation study to examine (i) sampler convergence behaviour across datasets with different statistical characteristics, and (ii) the effect of the initialisation strategy by comparing tBayes-MICE-V1 (mean-based initialisation) and tBayes-MICE-V2 (time-aware initialisation). 

We first examine convergence behaviour to assess the robustness of tBayes-MICE across datasets. The diagnostics in \cref{tab:convergence_summary} for the HCO\textsubscript{3} variable using the PhysioNet dataset reveal that all model parameters have $\hat{R}$ values close to 1 and large effective sample sizes under both MCMC samplers, indicating robust posterior exploration. However, as shown in \cref{tab:ablation_convergence_summary}, the CO(GT) variable from the AirQuality dataset demonstrates satisfactory convergence across all parameters displayed, with $\hat{R}$ values remaining below the accepted threshold of $1.05$. While a few regression coefficients (e.g., $w_2$ and $w_3$) exhibit lower ESS and slightly elevated $\hat{R}$ values (more than 1.05), although most of the key parameter values still fall within acceptable limits, indicating adequate mixing and convergence for these two parameters. This behaviour is likely due to the higher volatility, multicollinearity, and nonstationarity characteristics of environmental time series data, which can affect mixing for certain parameters. It is worth noting that the model achieves stable posterior summaries and good predictive performance despite this, suggesting that tBayes-MICE provides reliable imputations while appropriately capturing uncertainty in parameter estimates.

We next compare the two tBayes-MICE variants under the RWM sampler to assess the impact of the initialisation strategy (\cref{tab:metrics_rwm_datasets}). On the AirQuality dataset, both variants achieve nearly identical performance, with only marginal differences across variables, suggesting that time-aware initialisation does not degrade predictive accuracy. On the PhysioNet dataset, similarly small but consistent improvements are observed for temporally coherent variables such as HCO\textsubscript{3} and Na under tBayes-MICE-V2, suggesting time-aware initialisation improves robustness in smoother clinical signals. Overall, these results indicate that the primary gain stems from Bayesian formulation, while the time-aware initialisation in tBayes-MICE-V2 provides additional stability for temporally structured variables and across datasets. Accordingly, we adopt tBayes-MICE-V2 as the default configuration in this study.

\begin{table*}[t]
\centering
\caption{Convergence summaries for the CO(GT) variable from the AirQuality dataset using tBayes-MICE. Results are shown for two MCMC samplers: MALA (left) and RWM (right). The table reports posterior means, standard deviations (SD), 94\% highest density intervals (HDIs), effective sample sizes (ESS), and $\hat{R}$ statistics.}
\begin{subtable}{0.49\textwidth}
\centering
\caption{MALA--MCMC}
\scriptsize
\setlength{\tabcolsep}{4pt}
\renewcommand{\arraystretch}{1.15}
\resizebox{\linewidth}{!}{%
\begin{tabular}{
l
S[table-format=1.3]
S[table-format=1.3]
S[table-format=2.3]
S[table-format=2.3]
S[table-format=5.0]
S[table-format=1.2]
}
\toprule
{Param} & {Mean} & {SD} & {HDI$_{3\%}$} & {HDI$_{97\%}$} & {ESS} & {$\hat{R}$} \\
\midrule
$b$    &  0.000 & 0.006 & -0.012 &  0.011 & 4799 & 1.00 \\
$\tau$ &  0.026 & 0.001 &  0.023 &  0.028 & 38779 & 1.00 \\
RMSE   &  0.156 & 0.001 &  0.155 &  0.157 &  476 & 1.01 \\
\midrule
$w_1$  &  0.063 & 0.016 & 0.032 &  0.093 &  182 & 1.02 \\
$w_2$  &  0.961 & 0.048 &  0.874 &  1.047 &  14 & 1.12 \\
$w_3$  &  -0.279 & 0.047 &  -0.360 &  -0.189 &  11 & 1.13 \\
$w_4$  & -0.092 & 0.008 & -0.106 & -0.078 &  751 & 1.00 \\
$w_5$  &  0.061 & 0.007 &  0.048 &  0.075 &  1514 & 1.01 \\
$w_6$  &  0.036 & 0.006 & 0.024 &  0.048 &  3016 & 1.00 \\
\bottomrule
\end{tabular}}
\end{subtable}
\hfill
\begin{subtable}{0.49\textwidth}
\centering
\caption{RWM--MCMC}
\scriptsize
\setlength{\tabcolsep}{4pt}
\renewcommand{\arraystretch}{1.15}
\resizebox{\linewidth}{!}{%
\begin{tabular}{
l
S[table-format=1.3]
S[table-format=1.3]
S[table-format=2.3]
S[table-format=2.3]
S[table-format=5.0]
S[table-format=1.2]
}
\toprule
{Param} & {Mean} & {SD} & {HDI$_{3\%}$} & {HDI$_{97\%}$} & {ESS} & {$\hat{R}$} \\
\midrule
$b$    & 0.000 & 0.006 & -0.012 &  0.012 &  1384 & 1.00 \\
$\tau$ &  0.026 & 0.001 &  0.023 &  0.028 & 224 & 1.00 \\
RMSE   &  0.156 & 0.001 &  0.155 &  0.157 &  197 & 1.00 \\
\midrule
$w_1$  &  0.062 & 0.017 & 0.031 & 0.093 &  67 & 1.01 \\
$w_2$  &  0.967 & 0.054 & 0.866 &  1.054 &  29 & 1.05 \\
$w_3$  &  -0.289 & 0.048 & -0.378 & -0.204 &  38 & 1.02 \\
$w_4$  & `-0.091 & 0.008 & -0.106 & -0.077 & 437 & 1.00 \\
$w_5$  &  0.061 & 0.008 & 0.047 & 0.076 &  692 & 1.00 \\
$w_6$  &  0.036 & 0.006 & 0.025 & 0.049 &  905 & 1.00 \\
\bottomrule
\end{tabular}}
\end{subtable}
\label{tab:ablation_convergence_summary}
\end{table*}

\begin{table*}[!htbp]
\centering
\caption{Ablation study comparing tBayes-MICE-V1 and tBayes-MICE-V2 on the AirQuality and PhysioNet datasets under the RWM sampler. Metrics are NMAE and NRMSE (mean, standard deviation) across 30 runs. Boldface indicates the better-performing variant for each variable within each dataset.}
\small
\setlength{\tabcolsep}{6pt}
\renewcommand{\arraystretch}{1.15}
\begin{tabular}{lcccccc}
\toprule
 & \multicolumn{3}{c}{\textbf{AirQuality}} 
 & \multicolumn{3}{c}{\textbf{PhysioNet}} \\
\cmidrule(lr){2-4} \cmidrule(lr){5-7}
\textbf{Method}
& \textbf{Variable} & \textbf{NMAE} & \textbf{NRMSE}
& \textbf{Variable} & \textbf{NMAE} & \textbf{NRMSE} \\

\midrule
\multirow{4}{*}{tBayes-MICE-V1}
 & CO(GT)   & \textbf{0.1346 (0.0011}) & \textbf{0.1783 (0.0016)}
 & HCO\textsubscript{3}    & 0.7196 (0.0020 )& 0.9381 (0.0024) \\
 & NMHC(GT) & 0.2209 (0.0025) & 0.3464 (0.0029)
 & Na       & 0.7734 (0.0019) & 1.0263 (0.0028) \\
 &  C\textsubscript{6}H\textsubscript{6}(GT) & 0.1058 (0.0005) & \textbf{0.1556 (0.0015)}
 & Platelet & \textbf{0.6585 (0.0021)} & \textbf{0.9443 (0.0021)} \\
 & T        & \textbf{0.1345 (0.0016)} & \textbf{0.1859 (0.0018)}
 & WBC      & \textbf{0.6200 (0.0024)} & \textbf{0.9014 (0.0022)} \\

\midrule
\multirow{4}{*}{tBayes-MICE-V2}
 & CO(GT)   & 0.1349 (0.0013) & 0.1788 (0.0017)
 & HCO\textsubscript{3}     & \textbf{0.7189 0.0018)} & \textbf{0.9375 (0.0021)} \\
 & NMHC(GT) & \textbf{0.2201 (0.0021)} & \textbf{0.3449 (0.0030)}
 & Na       & \textbf{0.7724 (0.0024)} & \textbf{1.0246 (0.0032)} \\
 & C\textsubscript{6}H\textsubscript{6}(GT) & \textbf{0.1053 (0.0005)} & 0.1560 (0.0011)
 & Platelet & 0.6596 (0.0019) & 0.9465 (0.0018) \\
 & T        & 0.1383 (0.0013) & 0.1932 (0.0014)
 & WBC      & 0.6228 (0.0022) & 0.9039 (0.0026) \\

\bottomrule
\end{tabular}
\label{tab:metrics_rwm_datasets}
\end{table*}

\section{Discussion}

Our study introduces \emph{tBayes-MICE}, a Bayesian enhancement of the Multiple Imputation by Chained Equation framework. We assessed the model's Bayesian convergence using standard diagnostic criteria by Gelman et al. \cite{gelman1992inference} and compared its performance with baseline methods (MICE and BRITS) across two real-world datasets: an environmental time series (AirQuality) and a clinical time series (PhysioNet), using NMAE and NRMSE. We further conducted an ablation study to examine (i) sampler convergence behaviour across datasets with distinct statistical characteristics and (ii) the effect of the initialisation strategy by comparing two tBayes-MICE variants: a mean-based initialisation version (tBayes-MICE-V1) and a time-aware initialisation version (tBayes-MICE-V2). 

Across both datasets, our proposed tBayes-MICE achieved lower NMAE and NRMSE than Standard MICE and BRITS, and showed smaller variability across 30 experimental runs, suggesting improved stability. Although BRITS was competitive with some physiological variables (platelets and WBC) and occasionally achieved lower NMAE, the Bayesian framework provided more consistent overall performance. Nonetheless, the improvements are most pronounced in the clinical data, where preserving temporal structure is critical, highlighting the practical value of incorporating time-aware initialisation within the Bayesian MICE framework for longitudinal biomedical data.

We note that the uncertainty quantification from our results further differentiates the proposed Bayesian approach. The 95\% credible intervals adapt meaningfully over time, reflecting observed patterns in the training phase and widening appropriately in the test phase, especially in regions with limited information and rapid fluctuations. For CO(GT), we found that the increase in posterior uncertainty during test was modest, suggesting robust generalisation to unseen environmental dynamics. The Na variable exhibited wider intervals overall, reflecting a longer forecast horizon, greater inherent variability, and clinical noise in physiological measurements. Crucially, we note that the predictive means remained close to the true signals in both datasets, indicating that the model remains accurate even under uncertainty. Visual diagnostics, such as predicted-versus-true value scatter plots, residual scatter plots, sorted-index comparisons, and absolute-error box plots, confirmed these trends. Classical MICE and BRITS showed systematic bias, notably underestimating extreme values and exhibiting negative residual drift at elevated concentrations, while tBayes-MICE showed minimal bias and stable variance. Imputation error plots further confirm these findings, showing that the tBayes-MICE model exhibits error trajectories that are highly concentrated around zero, with relatively small fluctuations across missing indices, even under substantial missingness.

Another crucial practical insight from our study is that incorporating optimal step-size scaling recommendations from Robert et al. \cite{gelman1996efficient, gelman1997weak} for the MCMC proposal distribution significantly enhanced sampler performance and convergence stability. We note that adjusting the proposal variance in line with optimality theory (i.e., keeping acceptance rates close to theoretical optima, as recommended by Roberts and Rosenthal\cite{roberts2001optimal} for Metropolis methods) led the chains to mix more effectively and avoid excessively cautious or diffuse sampling patterns. This helped posterior trajectories stabilise more quickly and reduced the autocorrelation in the samples, leading to more reliable imputations. This scaling technique works well for all clinical variables and some environmental variables used in the study, underscoring the importance of providing theoretical guidance when tuning large Bayesian models with missing data.

In addition to improvements in overall accuracy, we notice significant differences in the sampling dynamics between the RWM and MALA samplers. Although both techniques yield almost identical prediction and imputation results, MALA demonstrated faster, smoother convergence and more efficient exploration of the posterior space for most variables. MALA exhibits reduced autocorrelation in the chains and stabilises at the posterior mode more rapidly, whereas RWM requires additional iterations to achieve similar mixing behaviour. Nonetheless, the similarity in the final predictive and imputation outcomes across samplers suggests that the tBayes-MICE framework is robust to sampler choice, with MALA offering faster convergence and higher sampling efficiency, while RWM remains a reliable baseline method.

Although the proposed tBayes-MICE framework performs well in terms of imputation accuracy, predictive quality, and uncertainty calibration across environmental and clinical time-series data, we identify some limitations. First, the evaluation was performed on two benchmark datasets, which, despite representing different domains, require additional validation on larger and more diverse time-series datasets (e.g., finance, mobility) to establish generalisability. In addition, the current experiments focus exclusively on continuous variables, and extending the framework to mixed-type data remains an important direction for future work. Secondly, although multivariate variables were imputed,  each variable was modelled independently, following MICE philosophy. A fully joint multivariate Bayesian time-series was not employed, meaning cross-feature temporal relationships were not captured within a single unified posterior. Extending this work to a multivariate Bayesian state-space \cite{ji2020bayesian, fang2023bayotide} or a Gaussian process model would allow the method to better leverage dependencies across variables. Thirdly, the MCMC sampling process incurs a higher computational cost than traditional MICE and BRITS (GPU usage), particularly for large feature sets, and careful tuning is also required.  Finally, the current evaluations assume missing-at-random (MAR) and missing-completely-at-random (MCAR) patterns. Future work will explore extensions of MICE to missing-not-at-random (MNAR) scenarios in spatio-temporal datasets.
Overall, our results provide strong evidence that a fully Bayesian formulation of MICE is both practical and beneficial, offering improved accuracy and uncertainty calibration for complex time-series imputation and laying the groundwork for scalable Bayesian imputation in high-dimensional real-world applications, with code made available on GitHub.

\section{Conclusion}
We presented the tBayes-MICE model, a Bayesian version of traditional MICE that explicitly accounts for uncertainty during imputation and improves handling of missing data in time-series analyses. By integrating MCMC sampling, the method not only improved imputation accuracy but also provided clearer, more principled uncertainty estimates via posterior credible intervals. We evaluated the proposed framework on two real-world datasets (AirQuality and PhysioNet) using NMAE and NRMSE, where tBayes-MICE consistently outperformed standard MICE and BRITS across most variables.  

An important observation from the ablation study is that the two tBayes-MICE variants exhibited nearly identical residual behaviour and predictive accuracy across experiments, indicating that the proposed Bayesian framework is robust to the choice of initialisation strategy. This stability suggests that performance gains arise from the Bayesian formulation itself rather than from specific design choices. Regarding the sampling schemes, both RWM and MALA achieved comparable predictive accuracy, as measured by the metrics; however, MALA demonstrated faster convergence and more efficient exploration of the posterior distribution. 

Overall, the tBayes-MICE framework has the potential to provide greater reliability and interpretability for imputing time-dependent data than conventional methods, which either neglect temporal relationships or are incapable of representing uncertainty. The results from our research encourage further development to extend the model to a fully multivariate Bayesian time-series framework for spatio-temporal datasets; to explore alternative sampling methods, such as Hamiltonian Monte Carlo or variational approximations; to incorporate non-linear relationships; and to provide comparative analysis with additional deep-learning-based methods.

\section*{Authors Contributions}
\textbf{Amuche Ibenegbu}: Conceptualisation; methodology; coding; data curation, analysis and experiment; writing - original draft. \textbf{Pierre Lafaye de Micheaux:} Conceptualisation, supervision, analysis and experiment, writing, review and editing.
\textbf{Rohitash Chandra:} Conceptualisation, supervision, analysis and experiment, writing - review and editing.
\section*{Declarations}

\begin{itemize}

\item \textbf{Conflict of interest}: The authors declare no competing interests

\item \textbf{Consent for publication}: All authors have read the manuscript and have given consent for submission and publication.

\end{itemize}

\section*{Availability of data}
The data and code can be accessed at the associated GitHub repository: \url{https://github.com/sydney-machine-learning/Bayes_MICE}

\section*{Acknowledgements}
The authors acknowledge the Katana High Performance Computing (HPC) cluster from the University of New South Wales. 

\nocite{*}
\bibliographystyle{plain}
\bibliography{refs}

\end{document}